\documentclass[10pt,twocolumn,letterpaper]{article}

\usepackage[pagenumbers]{wacv} 

\usepackage{graphicx}
\usepackage{amsmath}
\usepackage{amssymb}
\usepackage{booktabs}
\usepackage{multirow}
\usepackage{array}
\usepackage{wrapfig,lipsum,booktabs}
\usepackage[dvipsnames]{xcolor}
\usepackage{hyperref}
\usepackage{url}
\usepackage{soul}

\usepackage{algorithm}
\usepackage{algorithmic}
\usepackage{enumitem}

\newcommand{\simar}[1]{{\leavevmode\color{blue}#1}}

\usepackage[capitalize]{cleveref}
\crefname{section}{Sec.}{Secs.}
\Crefname{section}{Section}{Sections}
\Crefname{table}{Table}{Tables}
\crefname{table}{Tab.}{Tabs.}

\def\wacvPaperID{1534} 
\def\confName{WACV}
\def\confYear{2025}

\begin{document}

\title{Towards Secure and Usable 3D Assets: A Novel Framework for Automatic Visible Watermarking}

\author{Gursimran Singh, Tianxi Hu$^*$, Mohammad Akbari$^*$, Qiang Tang, Yong Zhang\\
Huawei Technologies Canada Co. Ltd.\\
{\tt\small \{gursimran.singh1, cindy.hu1, mohammad.akbari, qiang.tang, yong.zhang3\}@huawei.com}
}

\maketitle

\def\thefootnote{*}\footnotetext{Equal contribution.}
\def\thefootnote{\arabic{footnote}}

\begin{abstract}
3D models, particularly AI-generated ones, have witnessed a recent surge across various industries such as entertainment. Hence, there is an alarming need to protect the intellectual property and avoid the misuse of these valuable assets. As a viable solution to address these concerns, we rigorously define the novel task of automated 3D visible watermarking in terms of two competing aspects: watermark quality and asset utility. Moreover, we propose a method of embedding visible watermarks that automatically determines the right location, orientation, and number of watermarks to be placed on arbitrary 3D assets for high watermark quality and asset utility. Our method is based on a novel rigid-body optimization that uses back-propagation to automatically learn transforms for ideal watermark placement. In addition, we propose a novel curvature-matching method for fusing the watermark into the 3D model that further improves readability and security. Finally, we provide a detailed experimental analysis on two benchmark 3D datasets validating the superior performance of our approach in comparison to baselines. Code and demo are available \href{https://developer.huaweicloud.com/develop/aigallery/notebook/detail?id=15adbaaa-2583-4ec3-804a-61c29f001e03}{here}\footnote{\url{https://developer.huaweicloud.com/develop/aigallery/notebook/detail?id=15adbaaa-2583-4ec3-804a-61c29f001e03}}.
\end{abstract}
\section{Introduction}
\label{sec:intro}

The increasing demand for 3D assets across industries like entertainment \cite{liberatore_virtual_2021,OHailey_animation_2010},  augmented reality, and virtual reality \cite{parekh_systematic_2020,venkatesan_virtual_2021} has driven the adoption of efficient, and scalable methods for creating and distributing content. Generative AI (GenAI) has revolutionized automated 3D content creation, while commercial marketplaces have enhanced distribution networks. 
This evolution necessitates robust mechanisms to prevent misuse, validate ownership, and protect intellectual property (IP).

Governments and law enforcement agencies are concerned about potential misuse by malicious actors who may exploit automated AI tools to produce 
controversial 3D content at scale.
Such content could be used for spreading misinformation, influencing public opinion, or provoking social unrest. In response, governments (USA \cite{house_executive_2023}, China \cite{china_genai_2023}, and Europe \cite{noauthor_eu_2023}) are exploring regulations that would mandate GenAI services to embed and publicly disclose the origin of their generated content. 
In addition, in 3D data marketplaces, sellers must present their 3D assets for potential customers to preview using built-in 3D viewers. However, this practice can be exploited by malicious individuals who download these assets under the guise of previewing them, resulting in significant financial losses for the merchants. 
Therefore, safeguarding the intellectual property (IP) of these valuable assets within 3D data marketplaces is crucial to prevent unauthorized distribution and ensure fair compensation for the creators.


Digital watermarking is a key technology for copyright protection, source tracking, and authentication \cite{ranjbar2023nft,alvar2024amuse,rezaei2024lawa}. The majority of existing watermarking work in 3D focuses on invisible methods \cite{benedens_geometry-based_1999,bors_optimized_2013,uccheddu_wavelet-based_2004,wang_invisible_2020,yoo_deep_2022}. However, such methods have several drawbacks that limit their practicality in the scenarios discussed above. 
For instance, in the regulation scenario, it is essential for the general public to identify the asset's origin visually, without requiring specialized tools or knowledge \cite{house_executive_2023 ,china_genai_2023}. 
In contrast, invisible methods rely on specific, often publicly inaccessible, extractors that complicates watermark detection for non-experts. Furthermore, when 3D models are employed in downstream applications like video games or animations, the original watermarked assets become inaccessible for extraction. Consequently, extraction must rely on analyzing 2D visuals that undergo significant alterations, such as changes in lighting and texture, which can hinder successful extraction \cite{yoo2022deep}.
Lastly, invisible watermarks, designed to be hidden and imperceptible, do not sufficiently deter unauthorized use in a preemptive manner. 
Instead, they are more suited as a remedy after the alleged infringement has already happened.
In the digital marketplace, copyright infringement can be easily committed through the replication of digital assets. However, identifying and proving such infringement is challenging, leading to complex and costly legal actions.

In this work, we propose the novel task of automated 3D visible watermarking as a viable alternative.
The objective is to embed a copyright mark, such as a visible logo or text message, across different areas of the 3D model surface. These visible watermarks are designed to overcome the limitations associated with invisible methods discussed previously. 
Specifically, in regulatory contexts, smaller visible watermarks can be strategically positioned in less intrusive areas of 3D models. This approach aims to maintain the model's functionality while using watermarks as stamps to track provenance information. Thus, the general public can readily interpret visible watermarks to discern whether the content is AI-generated or created by humans.
On the other hand, for merchandising scenarios, we can strategically place large and prominent visible watermarks at various locations on the asset that act as a strong deterrent against unauthorized use.
This encourages interested customers to purchase the unwatermarked model, ensuring fair compensation for the asset owner. Hence, this approach offers proactive protection of intellectual property compared to relying solely on legal action after misuse. 


Automated 3D visible watermarking is a challenging task due to several reasons. First, the algorithm needs to automatically identify the most suitable locations (specific coordinates) on the 3D model surface to embed watermarks. In other words, the algorithm must choose locations that boost the visibility of watermarks for high security, while preserving the most salient and prominent features of the original model. Second, the algorithm must guarantee that the entire watermark is well-aligned along the model’s surface curvature with no 
component floating, as such parts can be easily removed using an isolated parts detection algorithm, thereby compromising the security of the watermarks. Finally, the algorithm must fuse the watermarks into the 3D geometry rather than merely inscribing them into the model texture. This makes it an irreversible process where the watermarks are very hard and expensive to remove. 

To address these challenges, we introduce an end-to-end pipeline for automatically embedding visible watermarks on arbitrary 3D models. Our solution proposes a novel gradient-based optimization method that generates a large number of candidate watermarks placed throughout the surface of the model. Having generated these diverse candidates, we define a filtering-based algorithm to pick the final set of watermarks based on the specific utility and security requirements. Finally, we propose a novel Boolean-based mesh merging to fuse watermarks into the geometry while matching the local curvature of the underlying surfaces. The major contributions of this work are as follows: 
\begin{itemize}[leftmargin=*]
    \item We define the novel task of automatic 3D visible watermarking, where the main goal is to determine the number, orientation, and location of watermarks to achieve the best watermark quality and best asset quality. We also provide rigorous definitions for various aspects of both watermark quality and asset quality. 
    \item We propose a novel end-to-end pipeline as a solution for the task of automated 3D visible watermarking. Our solution is based on a gradient-based optimization to determine the location and rotation transforms for the best orientation. Additionally, we propose a novel curve-matching fusion for enforcing the watermarks to follow the local curvature of the underlying surface. 
    \item We propose a holistic evaluation benchmark for 3D visible watermarking for future research. We propose practical metrics for various aspects of watermark and asset quality. Further, we provide detailed experiments on three 3D datasets to demonstrate the superiority of our approach in comparison to the baselines.
\end{itemize}

\simar{

}

\vspace{-5mm}
\section{Related Works}
\label{sec:related}

Watermarking 3D models is a problem of great interest \cite{an_visible_2016, benedens_geometry-based_1999, bors_optimized_2013, cayre_application_2003, yoo_deep_2022}. 
While 3D invisible watermarking \cite{benedens_geometry-based_1999, bors_optimized_2013, uccheddu_wavelet-based_2004, wang_invisible_2020, yoo_deep_2022} is well-established, the field of 3D visible watermarking remains in its infancy. Traditional approaches \cite{an_visible_2016, cao_reversible_2020, peng_visible_2021} use edge subdivisions to carve watermark characters on a smooth surface of the mesh. These approaches give a visual effect of watermark characters, however, they are designed to be only viewed using mesh-editing software that can display edge information. Hence, they are not visible to the naked eye in rendering mode or after printing the 3D model. Another limitation of these methods is that they do not support 3D formats without any edge information (e.g., voxel format). Recently, Li et 
al. \cite{li_visible_2024} proposed an approach to embed a single 3D watermark on a model using mesh Boolean operations, a technique similar to ours. However, their method has several limitations. First, they do not provide a framework for automatic identification of the locations to put watermarks on the surface of the model. Instead, they only provide some guidelines for humans to select locations manually. Second, their algorithm does not support fixing the orientation of watermarks, which is critical to ensure security against removal attacks. Finally, the specific Boolean merging operations proposed in this work do not match the curvature of the underlying surface, leading to poor visual quality of results. On the other hand, our method automatically selects and places watermarks on the surface of the 3D mesh with a focus on both watermark and asset quality.

\section{Problem Definition}
\label{sec:problem}
At a high level, the task of 3D visible watermarking is to automatically generate the watermarked (output) model by embedding multiple watermarks on the target (input) model. Formally, the inputs to the task include: 1) \textit{Target 3D mesh} $M(V, F)$ consisting of a set of vertices $V = \{v_i | v_i \in \mathbb{R}^3\}$, a set of faces $F = \{(v_i, v_j, v_k) | v_i, v_j, v_k \in V\}$; 2) watermark text; and 3) algorithm parameters such as font, thickness, and size of watermarks.

The output is \textit{a 3D mesh} $M'(V', F')$ with a variable number of watermarks, denoted by $H_f$,  embedded into its surface at multiple locations. The new set of vertices $V'$ and faces $F'$ represent the modified geometry where the watermarks are an inseparable part of the watermarked mesh $M'$. Additionally, let’s denote the sub mesh corresponding to the $i^{th}$ embedded watermark as $W_i \subset M'$ and an assumed minimum volume bounding box around it by $B_i$.


The central problem in 3D visible watermarking is to determine the number, location, and orientation of the watermarks to be placed on the target model's surface. For instance, having too many watermarks can obscure crucial details of the model, making it unusable for downstream applications. Conversely, too few watermarks may compromise security, allowing adversaries to exploit views without watermarks to create 2D renderings illegitimately. Similarly, the location and orientation of the watermarks have security-vs-utility consequences that need to be kept in mind during the watermarking process. To make it concrete, we group these security and utility consequences into two main groups - \textit{watermark quality} and \textit{asset utility}, and discuss them in detail in the following sub-sections.


\subsection{Watermark Quality}
\label{sec:watermark_quality}
Watermark quality encompasses various aspects of the security of the watermarked asset. In order to have high security, the placed watermarks must be 1) hard to remove by an adversary, 2) easily readable by a human eye, and 3) viewable from multiple camera angles. To make the problem concrete, we define these individual aspects into quantifiable and measurable metrics called watermark placement and watermark visibility.

\textbf{Watermark Placement: }
{The principle is that watermarks should be precisely aligned on the surface of the target mesh to enhance security, readability, and visual quality. Misaligned watermarks, where characters are embedded within the mesh surface, are undesirable for several reasons. Firstly, isolated characters can be easily detected and removed by automated algorithms, compromising the security of the watermark. Secondly, embedded parts make it difficult to fully read the watermark, affecting its readability. Lastly, watermarks that do not conform to the mesh surface appear visually less appealing to human observers.}

\textbf{Definition 1.} \textit{
{Watermark placement is quantified as the average proportion of the intersection area between the original mesh $M$ and each watermark mesh $W^i$ from the set of all watermarks $\{W^i\}_{i=1}^{H_f}$. This is computed as:}
}
\vspace{-4pt}
{\small\begin{equation}
\mathcal{P}(M, M') = \frac{1}{H_f}\sum_{i=1}^{H_f} \max \left \{ \frac{\mathcal{A}_{\hat{n}^i}(M \cap W^i)}{\mathcal{A}_{\hat{n}^i}(W^i)}, 1 \right \}
\label{eq:metricwaterplacement} 
\end{equation}
}

where $\mathcal{A}_{\hat{n}}(w)$ represents the projected area of a mesh $w$ towards a normal vector 
$\hat{n}$. $W^i \subset M'$ is the $i^{th}$ watermark mesh and $\hat{n}^i$ is the normal vector facing the 
front side of the corresponding watermark mesh.

\textbf{Watermark Visibility:} 
{It quantifies whether watermarks are clearly visible and readable from various camera angles around the watermarked mesh, which is crucial for enhancing security, as higher visibility prevents adversaries from obtaining renders (2D shots) of the asset without watermarks. Additionally, improved visibility ensures better readability, enabling investigators to easily identify and verify the presence of the watermark from multiple angles.}

\textbf{Definition 2.} \textit{
{Consider a set of camera views $\{c^t\}_{t=1}^T$ obtained by randomly rotating a camera around the watermarked object. Assume $\mathcal{K_V}(M', c^t)$ as a kernel, where it equals 1 if a watermark on $M'$ is visible in view $c^t$ to a human, and 0 otherwise. For a large number $T$ of views, we define the watermark visibility $V(M')$ as the proportion of views where at least one watermark is visible:}
}
\vspace{-4pt}
{\small\begin{equation}
\mathcal{V}(M') = \frac{1}{T}\sum_{t=1}^T \mathcal{K_V}(M', c^t) 
\label{eq:metricwatervisibility} 
\end{equation}}

\begin{figure*}[t]
  \centering
   \includegraphics[width=1\linewidth]{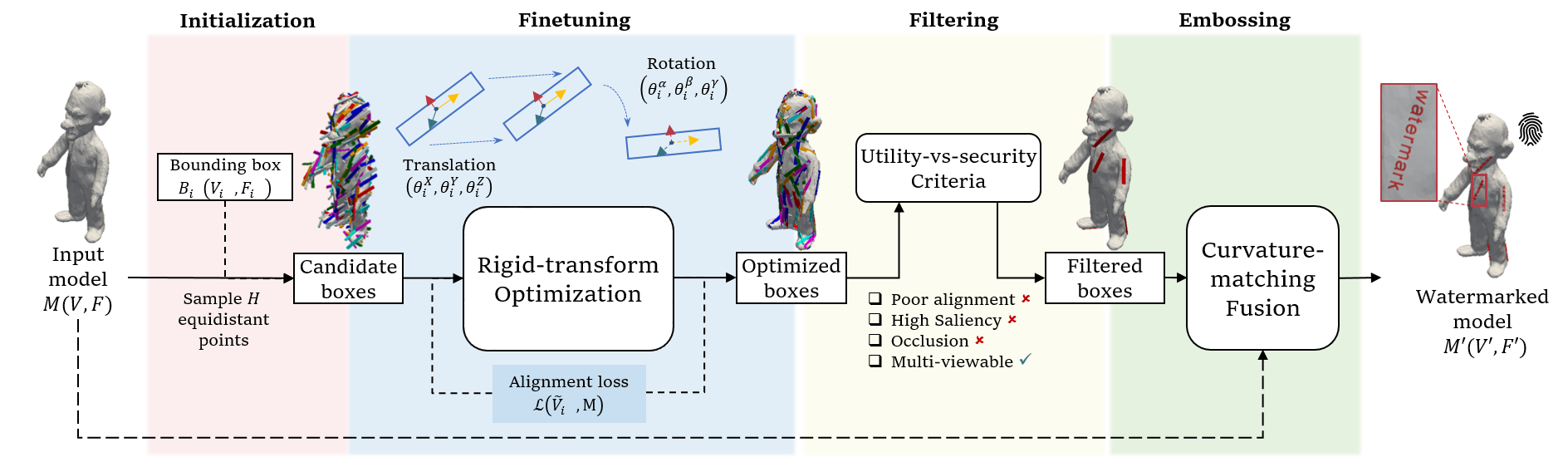}
   \vspace{-15pt}
   \caption{The overall framework of our proposed automatic 3D visible watermarking.}
   \label{fig:framework}
\end{figure*}

\subsection{Asset Utility}
\label{sec:asset_utility}
{Asset utility encompasses various factors that affect the usefulness of the watermarked asset across different downstream applications. Given the subjective nature and complexity of defining downstream utility, our focus is primarily on assessing any potential degradation caused by watermarking. Specifically, we evaluate how the watermarking process impacts the asset in terms of geometry, saliency, and semantic integrity.
For example, adding a greater number or larger size of watermarks to an asset results in more significant degradation. Similarly, placing watermarks in highly noticeable and salient areas of the asset can diminish its utility more than choosing less conspicuous, flat areas. Accordingly, we define specific criteria for assessing asset utility: geometry similarity, saliency preservation, and semantic preservation.}

\textbf{Geometry Similarity:} 
{This metric quantifies alterations in the appearance-level 3D geometry of the watermarked asset \( M' \) relative to the original asset \( M \). It focuses on detecting changes such as surface geometry modifications, variations in local curvature, and occlusions of surface features resulting from the watermarking process.}

\textbf{Definition 3.} \textit{
{Let $K_G(M', M, dr)$ denote a kernel quantifying the local 3D geometry similarity between a small region $dr$ of the target mesh $M$ and the corresponding region of the watermarked mesh $M'$. The geometry similarity is defined as:}
}
\vspace{-4pt}
{\small\begin{equation}
\mathcal{G}(M, M') = \int_{r} \mathcal{K_G}(M, M', dr) \cdot dr 
\label{eq:metricassetgeometry}
\end{equation}}

\textbf{Saliency Preservation:} 
{This criterion quantifies whether the salient and highly noticeable features of the original mesh are preserved during watermarking. For instance, in the context of a cat object, critical features like the face, ears, and paws must be preserved to ensure the continued usefulness of the model for downstream applications.}

\textbf{Definition 4.}  \textit{
{Let $\mathcal{K_S}(r)$ denote a kernel for computing an estimate of the average saliency of a local region $r$ in the original model $M$. Using a threshold $\tau_s$ determined by Otsu’s method \cite{xu_characteristic_2011}, we define the saliency retention as:}}
\vspace{-4pt}
{\small\begin{equation}
\mathcal{S}(M, M') = \frac{1}{H_f} \sum_{i=1}^{H_f} \mathbb{I}(\mathcal{K_S}(M \cap B^i) < \tau_s) 
\label{eq:metricassetsaliency}
\end{equation}}
where $\mathbb{I}(c)$ is an indicator function whose value is 1 if $c$ is true, otherwise 0.

\textbf{Semantics Preservation:} 
{This criterion addresses the potential degradation in high-level semantic concepts caused by watermarking. To maintain high semantics preservation, watermarks should avoid locations that could alter or obscure the semantic understanding of the object. For example, adding numerous watermarks to the face of an animal might make it challenging to discern whether it semantically represents a cat or a dog.}

\textbf{Definition 5.} \textit{
{Let $\{c_o^t\}_{t=1}^T$ and $\{c_w^t\}_{t=1}^T$  denote sets of corresponding camera views obtained by repeatedly and randomly rotating a camera around the original and watermarked objects, respectively. Assume $\mathcal{K_F}(c_o^t, c_w^t)$ as a kernel for estimating the semantic similarity between corresponding views $c_{o}^{t}$ and $c_{w}^{t}$. For a large value of $T$, semantics preservation is defined as:}
}
\vspace{-5pt}
{\small\begin{equation}
\mathcal{F}(M, M') = \frac{1}{T} \sum_{t=1}^T \mathcal{K_F}(c_o^t, c_w^t) \label{eq:metricassetsemantics}
\vspace{-5pt}
\end{equation}}

\subsection{Watermark Quality vs. Asset Utility}
\label{sec:vs}
The ultimate goal of the 3D visible watermarking task is to obtain high asset utility and watermarking quality. Ideally, the asset utility before and after watermarking should be approximately the same. At the same time, we would like high-quality watermarks that are visible from all angles to the human eye. However, watermarking often involves a trade-off between asset utility and watermark quality. The goal of this work is to improve the fundamental trade-off by improving both aspects of asset utility and watermark quality together.

\section{Method}
{
In this section, we outline our proposed solution for automated 3D visible watermarking. \cref{fig:framework} presents a high-level visualization of the entire process. Our pipeline is composed of four main modules: initialization, finetuning, filtering, and embossing. Initially, we generate a large number of candidate boxes, that serve as placeholders for the actual 3D watermarks, on the surface of the target model. Next, the finetuning module adjusts the position and orientation of these boxes to ensure they flow along the surface of the mesh rather than protruding out or into the surface. Then, the filtering module selects the final subset of watermarks with an aim to fulfill the high watermark quality and high asset utility requirements. Finally, the embossing module integrates the 3D text watermark into the target model’s surface, ensuring the watermarks conform to the local surface curvature for optimal visual appeal.

\subsection{Initialization}
\label{sec:candidate}


We begin by sampling a set of \( H \) equidistant points on the surface of the target model. At each sampled point \( i \), we create a 3D rectangular box denoted as \( B_i(V_i, F_i) \) and orient it to face along the corresponding surface normal. These boxes collectively serve as placeholders that will later be substituted with 3D text watermarks. For a detailed description of the initialization algorithm, please consult the supplementary materials.

\subsection{Finetuning}
\label{sec:bounding}
The boxes generated in the initialization phase need to be fine-tuned for optimal watermark placement. Specifically, they need to be moved to appropriate locations and rotated to properly align along the surface of the model. Since the required translation and rotation are unknown in advance, we propose a novel rigid-transform optimization approach to estimate the necessary transforms automatically. 


To accomplish this, we introduce rotation parameters \((\theta^\alpha_i, \theta^\beta_i, \theta^\gamma_i)\) and translation parameters \((\theta^X_i, \theta^Y_i, \theta^Z_i)\). Here, the rotation parameters specify the angles of rotation, while the translation parameters indicate the displacements along the X, Y, and Z axes, respectively. Using these parameters, we apply translation and rotation operations to each candidate box. For the \(i^{th}\) candidate box, the transformed vertices \(\tilde{V}_i\) are obtained as follows:
\vspace{-2pt}
{\small
\begin{equation}
\tilde{V}_i = T_{C^X_i, C^Y_i, C^Z_i} \cdot R_{\theta^\alpha_i, \theta^\beta_i, \theta^\gamma_i} \cdot -T_{C^X_i, C^Y_i, C^Z_i} \cdot \bar{V}_i 
\label{eq:optrot}
\end{equation}
\begin{equation}
\text{where } \bar{V}_i = T_{\theta^X_i, \theta^Y_i, \theta^Z_i} \cdot V_i, \label{eq:opttrans}
\end{equation}}
where \cref{eq:opttrans} represents the parameterized translation step of moving the vertices of $i^{th}$ bounding box $V_i$ using the translation matrix $T_{\theta^X_i, \theta^Y_i, \theta^Z_i} \in \mathbb{R}^{3 \times 3}$. \cref{eq:optrot} represents the parameterized rotation of the box around its centroid using the rotation matrix $R_{\theta^\alpha_i, \theta^\beta_i, \theta^\gamma_i} \in \mathbb{R}^{3 \times 3}$. Note, the translation matrix $-T_{C^X_i, C^Y_i, C^Z_i}$ is required to temporarily move the box at origin (0, 0, 0), which is eventually transported back to its previous place $(C^X_i, C^Y_i, C^Z_i)$ determined by \cref{eq:opttrans}. Moving the box temporarily to the origin is required to guarantee that the box is rotating around its centroid $(C^X_i, C^Y_i, C^Z_i)$ and not an arbitrary point.

Finally, we define the optimization objective as follows:
\vspace{-4pt}
{\small\begin{equation}
\label{eq:opt} 
\underset{{\{\theta^\alpha_i, \theta^\beta_i, \theta^\gamma_i, \theta^X_i, \theta^Y_i, \theta^Z_i\}_{i=1}^H}}{\arg\min} \frac{1}{H}\sum_{i=1}^{H} \mathcal{L}(\tilde{V}_i, M),
\end{equation}}
where $\mathcal{L}$ is the loss function defined in the next section.

\textbf{Loss Function.} In essence, it provides an assessment of the discrepancy in alignment between the box \(\tilde{V}_i\) and the mesh surface \(M\). Ideally, the mesh surface should intersect the box at its midpoint, effectively halving it. This alignment guarantees that the characters of the embedded watermark remain parallel to the surface, ensuring optimal readability and security.


Let \( t_1, t_2, t_3, t_4 \) denote consecutive vertices on the front face of the box, and \( b_1, b_2, b_3, b_4 \) denote corresponding vertices on the bottom face. This gives us midpoints \( m_1 = \frac{b_1 + t_1}{2}, m_2 = \frac{b_2 + t_2}{2}, m_3 = \frac{b_3 + t_3}{2}, \) and \( m_4 = \frac{b_4 + t_4}{2} \) on the lateral faces of the box. Then, we simply sample equidistant points along the line segments defined by points \( (m_1, m_2), (m_2, m_3), (m_3, m_4), \) and \( (m_4, m_1) \). We denote all sampled points, including the midpoints, by the set \( \{s^j_i\}_{j=1}^J \), where \( J \) is the total number of points per box. Using this set, we define the loss as the distance between these sampled points and the surface of the mesh:


 \vspace{-5pt}
{\small\begin{equation}
\mathcal{L}(\tilde{V}_i, M) = \frac{1}{J}\sum_{j=1}^{J} \mathcal{D}(s^j_i, M),
\label{eq:loss}
\end{equation}}
where \(\mathcal{D}\) denotes the built-in differential loss in PyTorch3D \cite{ravi_accelerating_2020} used to compute distances between point clouds and meshes. Intuitively, minimizing this loss ensures that all points \(\{s^j_i\}_{j=1}^J\) lie on the mesh surface. This alignment is achieved when the mesh surface passes through the middle of the box, parallel to both the front and back faces, thereby bisecting the box into halves.


\textbf{Learning.} 
The optimization objective specified in \cref{eq:opt} is nonlinear and non-convex. We simply use back-propagation to compute gradients of objective in \cref{eq:loss} and obtain the optimized parameters 
{\scriptsize$\{\overset{*}{\theta_i^\alpha}, \overset{*}{\theta_i^\beta}, \overset{*}{\theta_i^\gamma}\}_{i=1}^H$} and {\scriptsize$\{\overset{*}{\theta^X_i}, \overset{*}{\theta^Y_i}, \overset{*}{\theta^Z_i}\}_{i=1}^H$ }
by minimizing the objective in \cref{eq:loss}.
This process yields the refined bounding-box vertices \(\{V^*_i\}_{i=1}^H\) and their corresponding meshes \(\{B_i^*(V^*_i, F_i)\}_{i=1}^H\).


Due to the non-convex nature of the optimization, it is possible to obtain sub-optimal solutions, resulting in poorly aligned boxes. To address this, we start with a large number \( H \) of candidate bounding boxes distributed across the mesh's surface. Then, we apply heuristics to eliminate sub-optimal solutions, as detailed in the following section.

\subsection{Filtering}
\label{sec:filtering}
This module aims to select a subset of size $H_f \ll H$ from the optimized boxes obtained in the previous step. Guided by the utility-vs-security criteria outlined in \cref{sec:problem}, the goal is to ensure robust watermark security while preserving the utility of the watermarked asset with minimal degradation. Specifically, we do this by incrementally pruning sub-optimal, unnecessary, and redundant watermarks using a series of filtering steps.


We start by rejecting sub-optimal boxes that are poorly oriented (high loss) or are placed on highly salient regions \cite{cao_reversible_2020} of the mesh. Then, we remove boxes that have poor visibility due to potential occlusion by sub-parts of the original mesh using a ray casting approach. Next, we strategically choose the minimal set of watermarks to meet security criteria. Specifically, the model is divided into eight octants by segmenting it with X, Y, and Z planes. Subsequently, we employ a greedy method to choose one box per octant, aiming to maximize the spacing between selected boxes in the solution set. Following this, we search for locations for watermarks using fixed angle increments of 30° around the X and Z axes. New watermarks are added if no existing watermarks are already positioned at that angle. These two steps ensure that the watermarks are viewable from multiple angles for high watermark security. Please refer to the supplement for more details about these steps.



\subsection{Embossing}
\label{sec:embossing}
Having obtained the locations and orientations of the final set of $H_f$ candidate boxes, we generate the corresponding 3D-text watermark meshes using a standard text-to-3D algorithm \cite{narayanan_codetigerfont23d_2024}. 
However, these watermarks remain discrete and disconnected objects (individual 3D characters), which could be identified and removed by a knowledgeable attacker, thereby compromising security. Additionally, these watermarks create a flat textual surface on potentially curved surfaces, which can significantly degrade the visual quality (utility) of the watermarked asset (see \cref{curve-matching}).


We propose a novel method for enhancing watermark meshes by aligning them with the local curvature of the underlying surface through a curve-matching fusion. Initially, we compute the intersection between the target model and the 3D-text watermark meshes. Subsequently, we extrude the intersection result at all vertices in the direction of the bounding box normal by a consistent distance. This adjustment results in updated watermark meshes that conform to the local curvature, maintaining a fixed distance from the underlying surface. Finally, we apply Boolean operations such as union and difference, as described in \cite{li_visible_2024}, to achieve embossing or debossing effects. For a detailed algorithm, please refer to the supplementary materials.

}

\section{Experiments}
{In this section, we analyze and compare the performance of the proposed automated 3D visible watermarking method both quantitatively and qualitatively against the baselines.}

\begin{table*}[htbp]
\centering
\footnotesize
\setlength{\tabcolsep}{4pt} 
\setlength\extrarowheight{1pt}
\begin{tabular}{l|l|ccc|ccccc}
\toprule
Dataset & Method & \multicolumn{3}{c}{Watermark Quality} & \multicolumn{5}{c}{Asset Utility} \\
 & & \begin{tabular}[c]{@{}c@{}}WPS $\uparrow$\end{tabular} & \begin{tabular}[c]{@{}c@{}}Ray $\uparrow$\end{tabular} & \begin{tabular}[c]{@{}c@{}}OCR $\uparrow$\end{tabular} & \begin{tabular}[c]{@{}c@{}}SMSE $\downarrow$\end{tabular} & \begin{tabular}[c]{@{}c@{}}IPE $\downarrow$\end{tabular} & \begin{tabular}[c]{@{}c@{}}LCE $\downarrow$\end{tabular} & \begin{tabular}[c]{@{}c@{}}SE $\downarrow$\end{tabular} & \begin{tabular}[c]{@{}c@{}}SS $\uparrow$\end{tabular} \\
\hline
\multirow{2}{*}{Manifold40} & Li et al. & 0.280 & 0.706 & 0.301 & 0.006 & 20.525 & 0.022 & 0.102 & 0.795 \\
 & Ours & \textbf{0.484} & \textbf{0.920} & \textbf{0.369} & \textbf{0.002} & \textbf{0.575} & \textbf{0.000} & \textbf{0.093} & \textbf{0.832} \\
\hline
\multirow{2}{*}{Objaverse} & Li et al. & 0.203 & 0.588 & 0.201 & 0.009 & 19.400 & 0.038 & 0.168 & 0.792 \\
 & Ours & \textbf{0.420} & \textbf{0.788} & \textbf{0.355} & \textbf{0.002} & \textbf{4.820} & \textbf{0.002} & \textbf{0.021} & \textbf{0.811} \\
 \hline
\multirow{2}{*}{Meshy} & Li et al. & 0.207 & 0.837 & 0.362 & 0.0075 & 27.75 & 0.0293 & 0.146 & 0.822 \\
 & Ours & \textbf{0.411} & \textbf{0.993} & \textbf{0.526} & \textbf{0.0017} & \textbf{0.35} & \textbf{0.0006} & \textbf{0.020} & \textbf{0.856} \\
\bottomrule
\end{tabular}
\vspace{-6pt}
 \caption{Comparison results in terms of different watermark quality and asset quality metrics on Manifold40, ObjaVerse, and Meshy datasets. 
  $\uparrow$: higher is better. $\downarrow$: lower is better.}
  \label{tab:results}
\end{table*}

\textbf{Baseline Methods.} To the best of our knowledge, there is no previous work with the capability of automatically determining the locations of watermarks. Hence, none of the existing baselines is directly applicable to the task of  3D visible watermarking presented in this work. For the sake of comparison, we have re-purposed the most recent and capable baseline Li et al. \cite{li_visible_2024} and assumed random locations on the 3D mesh for placing watermarks in an automated manner. All other methods \cite{an_visible_2016,cao_reversible_2020,peng_visible_2021} have significant shortcomings (see \cref{sec:related}) which limit their applicability to the practical scenarios of 3D visible watermarking.

\textbf{Datasets and Implementation Details.} 
{We utilize three 3D datasets for our experiments. The first two datasets consist of 50 models randomly sampled from well-known benchmark datasets: Manifold40 \cite{hu_subdivision-based_2022} and ObjaVerse \cite{deitke_objaverse_2022}. The third dataset comprises 20 textured models obtained from the Meshy \cite{noauthor_meshy_nodate} text-to-3D generative AI service.}
Our method dynamically chooses the number of watermarks as determined by the filtering step (\cref{sec:filtering}). For a fair comparison, in all our experiments, we choose the same number of watermarks for the Li et al. baseline. 
More implementation details, running time analysis, and statistics of the datasets are given in the supplementary materials.

\textbf{Evaluation Metrics. } 
{For watermark quality, we define three metrics, namely \textit{WPS}, \textit{Ray}, and \textit{OCR}, based on the concepts of watermark placement and visibility defined in Section \ref{sec:problem}. WPS is computed using \cref{eq:metricwaterplacement}, except that we approximate the watermark mesh $W_i$ by its oriented bounding box $B_i$ to efficiently compute the intersection area. The Ray and OCR scores are computed by \cref{eq:metricwatervisibility} using a fixed number of camera views $T$ sampled around the $X$ and $Z$ axis. Moreover, for Ray, the kernel  $\mathcal{K_V}(M', c^t)$ is implemented via ray casting that approximates viewability by firing multiple rays from the front face of a watermark's bounding box and checks whether all of them can reach the camera uninterrupted. 
On the other hand, for OCR, we use a standard OCR method \cite{keras-ocr_2024} to approximate $\mathcal{K_V}(M', c^t)$, that outputs 1 if at least one watermark can be correctly recognized in the 2D render of a certain view $c^t$.

For asset utility, we define five different scores. The first three, Sampled Mean Squared Error (SMSE), Isolated Parts Error (IPE), and Local Curvature Error (LCE) are based on geometric similarity defined in \cref{eq:metricassetgeometry}. 
SMSE score is similar to the concept of Mean Squared Error (MSE), except that it is between two meshes instead of a vector of points. 
The IPE metric calculates the difference in total isolated parts between the watermarked mesh \( M' \) and the original mesh \( M \). This difference increases when watermarks are poorly aligned on the model surface, causing certain 3D letters to float above or below the surface. 
The LCE score assesses curvature preservation post-watermarking (discussed in \cref{sec:embossing}). It calculates the variance in distance from the mesh surface to the top of the watermark in watermarked areas. For curve-preserving watermarks, this distance remains constant, resulting in lower variance.

\begin{figure}[t]
\vspace{-2mm}
  \centering
   \includegraphics[width=1.0\linewidth]{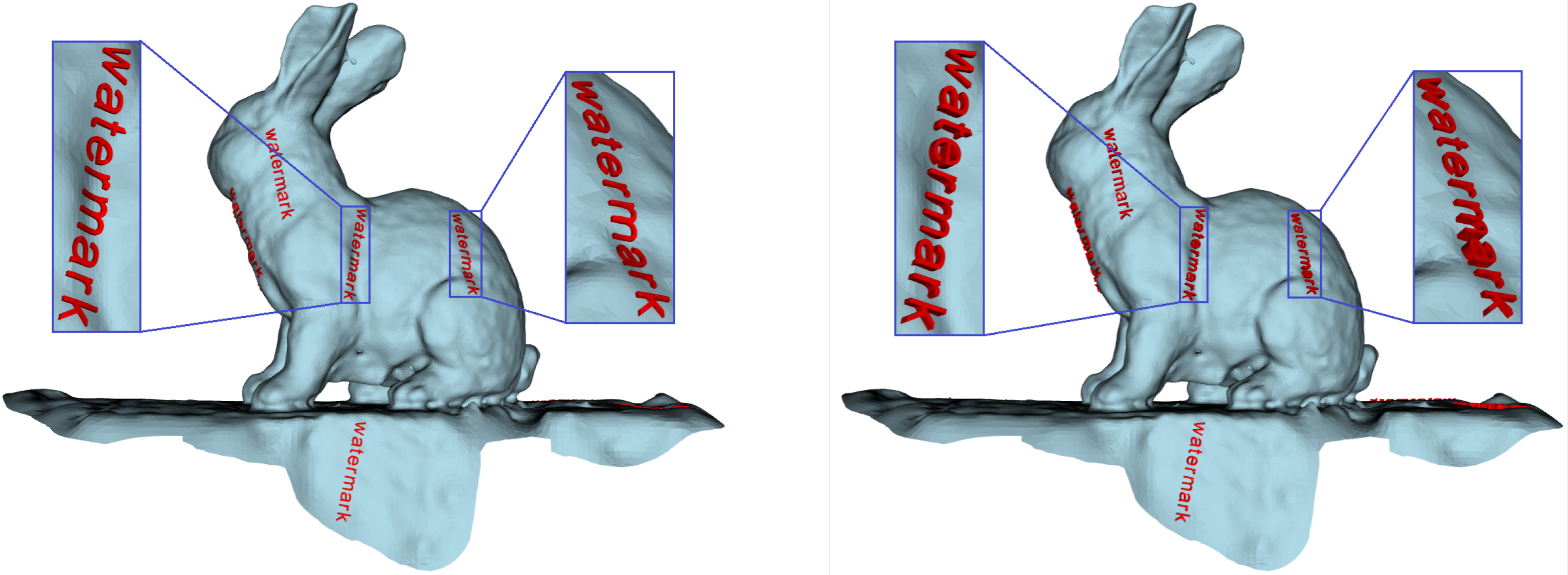}
   \vspace{-15pt}
   \caption{Qualitative analysis of our method with (left) and without (right) curve-matching fusion on an example from Objaverse.}
   \label{curve-matching}
   \vspace{-1mm}
\end{figure}

\begin{figure}[t]
  \centering
   \includegraphics[width=1.0\linewidth]{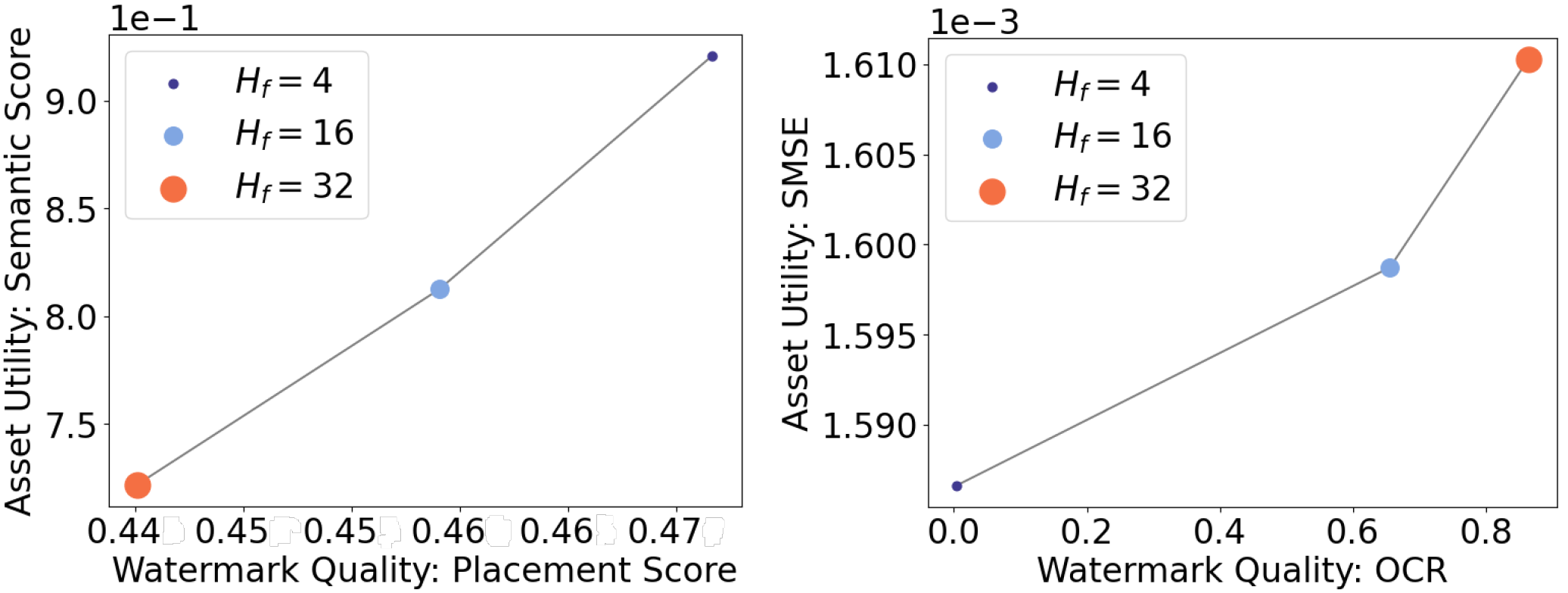}
   \vspace{-15pt}
   \caption{Trade-off results between the watermark quality and asset quality metrics on Manifold40. $H_f$: number of watermarks.}
   \label{fig:tradeoff}
\end{figure}

The remaining two scores Saliency Error (SE) and Semantic Score (SS) are based on definitions \cref{eq:metricassetsaliency} and \cref{eq:metricassetsemantics}. 
Specifically, to implement the saliency kernel $\mathcal{K_S}(w)$, we pick an off-the-shelf algorithm \cite{nousias_mesh_2020} to compute the saliency map and report the average saliency of the sub mesh $w$ as the output.
For the semantic kernel $\mathcal{K_F}(c_o^t, c_w^t)$, we compute the cosine similarity between the semantic features of 2D renders for views $c_o^t$ and $c_w^t$ using a pre-trained ResNet50 \cite{he_deep_2015} feature extractor.




}

\subsection{Quantitative Results}
\label{sec:quant}

The comparison results of our method and the baseline in terms of watermark quality and asset utility aspects are summarized in \cref{tab:results}. Overall, our approach outperforms the baseline on all the metrics by a significant amount. In particular, as indicated by the WPS and Ray scores, the watermarks embedded using our method achieve $\approx$20\% higher surface alignment (WPS) and multi-view visibility (Ray) compared to the baseline. Moreover, our OCR scores are 7\%, 15\%, and 16\% higher for Manifold40, ObjaVerse, and Meshy, respectively, which show the superiority of our method in watermark text readability.

On the other hand, for the asset utility, our method achieves $\approx$4X lower SMSE error rate on average, showing high geometry similarity between the watermarked and original 3D assets. In addition, we achieve significantly lower IPE and LCE error rates that guarantee very low isolated parts and high curvature preservation after watermarking, respectively. Further, the results also demonstrate our method’s superiority for the preservation of the salient features and semantic context of the asset after watermarking.

In order to study the trade-off between the watermark quality and asset utility (discussed in \cref{sec:vs}), we perform another set of experiments with a different number of watermarks $H_f = \{4, 16, 32\}$. Two trade-off curves including Semantic vs. Placement and SMSE vs. OCR are illustrated in \cref{fig:tradeoff}. Each point on the curve represents the results averaged over the models in the Manifold40 dataset. In general, increasing the number of watermarks results in higher watermark quality, but lower asset utility. However, compared to the baseline, our method achieves a significantly improved trade-off. In particular, for the same semantic score of $\approx$0.80 (given in \cref{tab:results}), our method achieves a placement score of 0.46 which is 18\% higher than the baseline. Moreover, our method can achieve an OCR score of 0.60, while providing an SMSE error of 0.0016. On the other hand, the baseline can achieve a much lower OCR score of 0.30 with a much higher SMSE error of 0.006. More trade-off curves are provided in the supplementary materials.

\subsection{Qualitative Results}
\label{sec:quali}
\begin{figure}[t]
    \vspace{-6mm}
  \centering
   \includegraphics[width=1.0\linewidth]{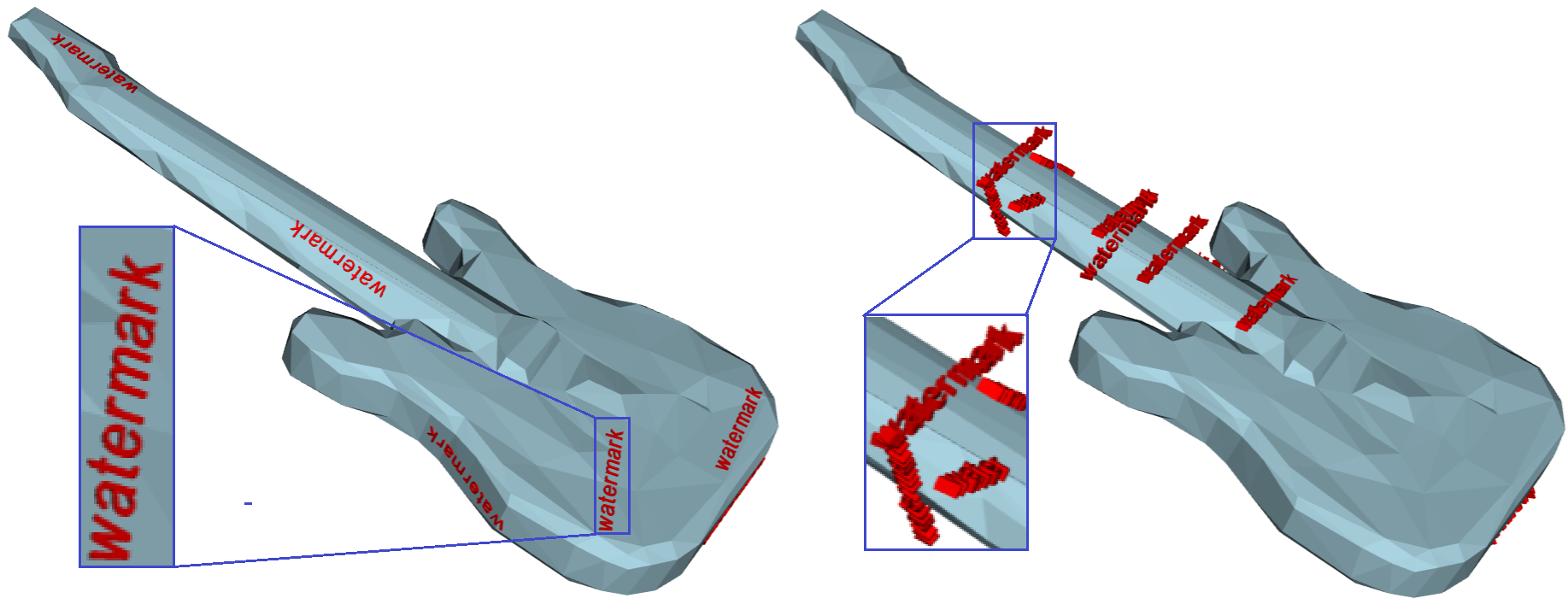}
   \\
   \vspace{0.5cm}
    \includegraphics[width=1.0\linewidth]{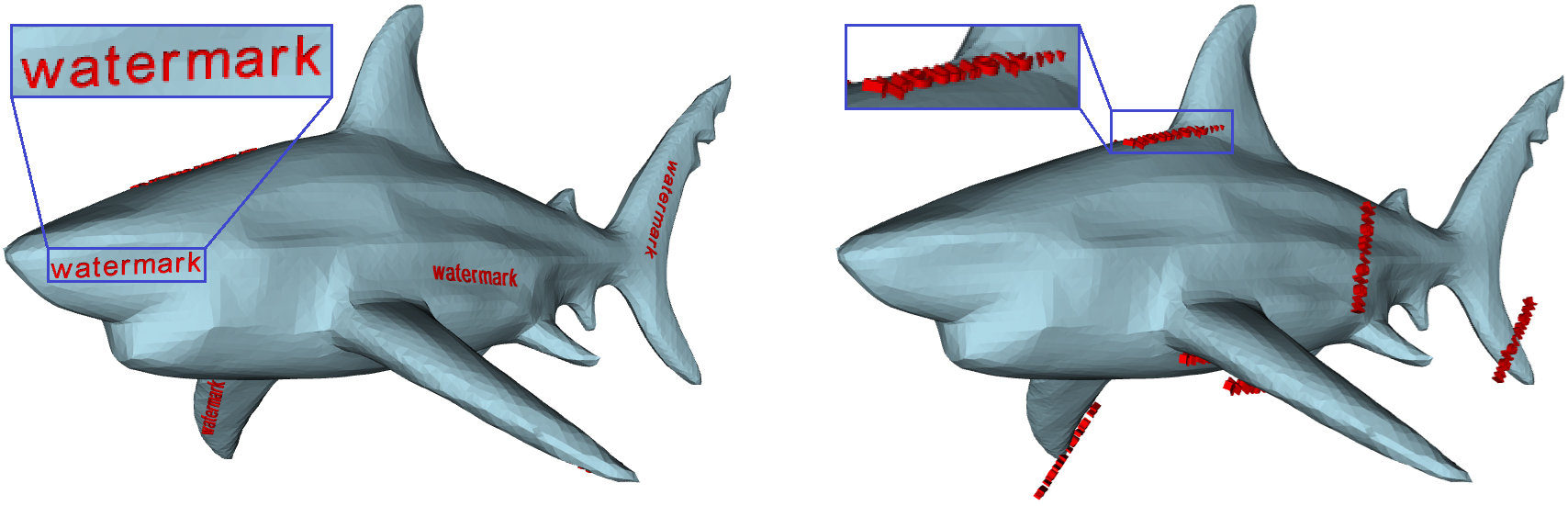}
    \vspace{-15pt}
   \caption{Visual example of 3D models from Manifold40 (top) and Meshy (bottom) watermarked with our method (left) and Li et al. baseline (right). Ours provides better placement quality, readability, and viewability.}
   \label{fig:compare}
\end{figure}

\cref{fig:compare} shows the qualitative results of our method and the baseline. The watermarks are colored red to enhance visibility for the reader. Our method generates watermarks that are better aligned with the surface and visible from multiple views. On the other hand, the baseline produces watermarks that either fly out or are hidden under the surface, 
resulting in poor visibility. 
Further, in \cref{curve-matching}, a qualitative analysis of our method with and without the proposed curve-matching fusion module is shown. From the enlarged areas, it is clear that including this fusion enforces the watermarks to follow the surface curvature, which provides higher watermark visibility and asset utility. More qualitative results along with analysis are given in the supplementary materials. Moreover, to subjectively analyze the performance of our method compared to the baseline, we conducted a user study, which is summarized in the supplementary materials.


\subsection{Ablation Study}
We perform an ablation study to analyze the effect of the 
components of our method including rigid-transform optimization (\cref{sec:bounding}), 
curvature fusion (\cref{sec:embossing}), and the filtering steps (\cref{sec:filtering}). The corresponding watermark quality and asset utility results on the ObjaVerse dataset are presented in \cref{tab:ablation}. As summarized in the table, excluding any of the optimization or filtering modules negatively affects both watermark quality and asset utility metrics. Specifically, our method without the optimization stage not only provides approximately 20\% lower WPS and Ray scores, but it also reduces the SMSE and LCE error rates by 2$\times$ and 7$\times$, respectively. For the filtering steps, the results are impacted less as the optimization module contributes more to the overall performance of the method. It is also seen that excluding the curve-matching fusion module results in a much higher LCE error, which shows the significance of this module in preserving the surface curvature of the asset.

\begin{table}[htb]
\centering
\scriptsize
\setlength{\tabcolsep}{2pt} 
\begin{tabular}{l|ccc|cccccc}
\toprule
Method & \multicolumn{3}{c}{Watermark Quality} & \multicolumn{4}{c}{Asset Utility} \\
 & \begin{tabular}[c]{@{}c@{}}WPS $\uparrow$\end{tabular} & Ray $\uparrow$ & OCR $\uparrow$ & SMSE $\downarrow$ & IPE $\downarrow$ & LCE $\downarrow$ & \begin{tabular}[c]{@{}c@{}}SS $\uparrow$\end{tabular} \\
\hline
Ours & \textbf{0.420} & \textbf{0.788} & \textbf{0.355} & \textbf{0.002} & 4.820 & \textbf{0.002} & \textbf{0.811} \\
\hspace{.1cm} \begin{tabular}[l]{@{}c@{}}w/o CF\end{tabular} & 0.420 & 0.788 & 0.294 & 0.002 & 4.820 & 0.021 & 0.781 \\
\hspace{.1cm} w/o Optimize & 0.225 & 0.543 & 0.232 & 0.005 & \textbf{3.800} & 0.015 & 0.786 \\
\hspace{.1cm} w/o Filtering & 0.287 & 0.594 & 0.259 & 0.002 & 7.870 & 0.004 & 0.800 \\
Li et al. & 0.203 & 0.588 & 0.201 & 0.009 & 19.400 & 0.038 & 0.792 \\
\bottomrule
\end{tabular}
\vspace{-4pt}
\caption{Ablation over different components of our method on the ObjaVerse dataset. CF: curvature fusion.}
\label{tab:ablation}
\end{table}

 \vspace{-3mm}
\subsection{Attack and Robustness Analysis}
\label{sec:attacks}


First, we analyze the robustness of our method in comparison to Li et al. \cite{li_visible_2024} baseline. We consider the cropping and unauthorized removal attacks, which are chosen to reflect strong adversaries. The results are summarized in \cref{tab:attacks}. In the crop attack, an attacker aims to illegitimately crop a significant part of the model. The attack severely impacts both watermark quality and asset utility. Despite this, our method surpasses the baseline as it applies watermarks at multiple angles, thereby preserving more watermarks and leading to improved watermark quality metrics. 

On the other hand, for the unauthorized removal attack, we assume the attacker can remove the vertices and faces belonging to the watermark, probably using manual or automated methods. As shown in \cref{tab:attacks}, our method preserves the watermark quality that degrades significantly for the baseline. This is because removing watermarks in our method leaves holes in the mesh surface, creating a silhouette through which the watermark can still be read clearly. In contrast, in baseline, parts of the watermark that are not in direct contact can be completely removed, resulting in the watermark being unreadable. Please refer to the supplement for a qualitative comparison.

\begin{table}[htbp]
\centering
\scriptsize 
\setlength{\tabcolsep}{4pt} 
\begin{tabular}{l|l|cc|cc}
\toprule
\multirow{2}{*}{Attack} & \multirow{2}{*}{Method} & \multicolumn{2}{c}{Watermark Quality} & \multicolumn{2}{c}{Asset Utility} \\ 
& & Ray $\uparrow$ & OCR $\uparrow$ & IPE $\downarrow$ & \begin{tabular}[c]{@{}c@{}}SS $\uparrow$\end{tabular} \\ \hline
\multirow{2}{*}{No Attack} & Li et al. & 0.706 & 0.301 & 20.525 & 0.795 \\ 
                           & Ours      & \textbf{0.920} & \textbf{0.369} & \textbf{0.575} & \textbf{0.832} \\ \hline
\multirow{2}{*}{Crop Attack}      & Li et al. & 0.501 & 0.093 & 9.650 & 0.685 \\ 
                           & Ours      & \textbf{0.569} & \textbf{0.120} & \textbf{4.050} & \textbf{0.716} \\ \hline
\multirow{2}{*}{Removal Attack}   & Li et al. & 0.541 & 0.211 & \textbf{20.350} & 0.918 \\ 
                           & Ours      & \textbf{0.920} & \textbf{0.356} & 29.025 & \textbf{0.927} \\ 
\bottomrule
\end{tabular}
\vspace{-6pt}
\caption{Quantitative results of our method compared to the baseline against crop and removal attacks.}
\label{tab:attacks}
\end{table}

\begin{figure}
  \centering
    \includegraphics[width=0.9\linewidth]{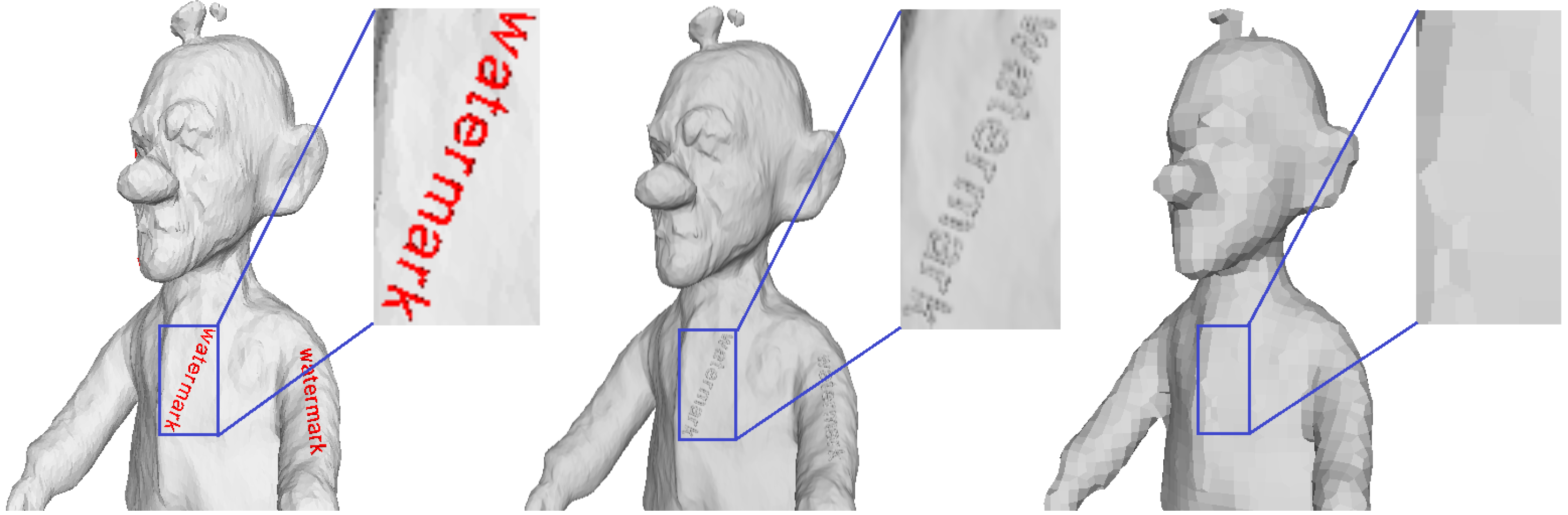}
    \label{fig:remeshing-attack}
    \vspace{-10pt}
   \caption{Left shows the original model; middle and right show the results of the remeshing attack with low and high strength.}   
\end{figure}

Lastly, we analyze the performance of our watermarks against the varying strengths of the remeshing attack, which is based on blender's Remesh Modifier \cite{blender/remesh}. As depicted in \cref{fig:remeshing-attack}, at low attack strength (middle), both asset utility and visible watermarks are reasonably preserved, albeit with loss of texture during remeshing. As the attack strength increases (right), visible watermarks are removed but the asset utility is also significantly impaired, particularly damaging facial features.

\setlength{\columnsep}{6pt}
\begin{wraptable}{r}{0.57\columnwidth} 
\vspace{-4pt}
\scriptsize
\setlength{\tabcolsep}{2.2pt} 
\begin{tabular}{l|cc|cc}
\toprule
Method & Ray $\uparrow$ & OCR $\uparrow$ & SMSE $\downarrow$ & SS $\uparrow$\\
\hline
Invisible & 0.000 & 0.000 & \textbf{7.81} $\times 10^{-9}$ & \textbf{0.999} \\ 
Ours & \textbf{0.920} & \textbf{0.369} & 0.002 & 0.832 \\
\bottomrule
\end{tabular}
\label{tab:invisible_quan}
\vspace{-7pt}
\caption{Comparison with invisible method, Wang et.al. \cite{wang_invisible_2020}.}
\vspace{-2pt}
\end{wraptable}

\subsection{Comparison with Invisible Watermarking}
\label{sec:comparision}
Here, we provide a comparison of performance and robustness with an invisible baseline, Wang et.al. \cite{wang_invisible_2020}. The performance results are shown in Tab. 4. As expected, the invisible technique performs poorly on the watermark visibility criteria of our benchmark, signaling its inability to meet the demands of the proposed task. Further, we found that this baseline was completely ineffective ($<50\%$ bit accuracy) against the cropping and remeshing attacks (even low strength), signaling its poor robustness in comparison to the proposed visible method. 

\section{Conclusion}
In this paper, we tackled the novel task of automatically embedding 3D visible watermarks to arbitrary 3D models. We first defined the objectives of the task of 3D visible watermarking in terms of various aspects of watermark and asset quality. Then, we proposed an end-to-end pipeline that uses a gradient-based optimization to achieve high watermark quality and high asset utility. We conducted an extensive set of experiments on two benchmark 3D datasets to demonstrate the effectiveness of our approach. Through our work, we aim to further research in the novel and practical direction of 3D visible watermarking. The limitations of our work are discussed in the supplementary materials.


\clearpage
\setlength{\topsep}{0pt}
\setlength{\partopsep}{0pt}
{\small
\bibliographystyle{ieee_fullname}
\bibliography{WACV}
}

\appendix
\section{Supplementary Materials}
In this appendix, we present the supplementary materials for the paper titled "\textit{Towards Secure and Usable 3D Assets: A Novel Framework for Automatic Visible Watermarking}".

\subsection{Code and Demo}
In order for the results to be reproducible, we share our test code with detailed instructions in the supplementary materials. We also uploaded a video
file demonstrating qualitative examples of watermarked 3D objects from various viewing angles. Code and demo are available \href{https://developer.huaweicloud.com/develop/aigallery/notebook/detail?id=15adbaaa-2583-4ec3-804a-61c29f001e03}{here}\footnote{\url{https://developer.huaweicloud.com/develop/aigallery/notebook/detail?id=15adbaaa-2583-4ec3-804a-61c29f001e03}}.

\subsection{Datasets Statistics}
As stated in Sec. 5, we sample a subset of 50 models from two benchmark 3D datasets, namely, Manifold40 and ObjaVerse. Particularly, we used random sampling stratified by output classes from the train set of the respective datasets. Additionally, for the Meshy GenAI dataset, we downloaded 20 textured models generated using the Meshy text-to-3D AI service. The statistics of the vertices and faces of these datasets are presented in \cref{tab:dataset_stats}.

\begin{table}[htbp]
\centering
\scriptsize 
\setlength{\tabcolsep}{2pt} 
\begin{tabular}{cc|ccc|ccc}
\toprule
\multicolumn{2}{c|}{\textbf{Dataset}} & \multicolumn{3}{c|}{\textbf{Vertices}} & \multicolumn{3}{c}{\textbf{Faces}} \\
\textbf{Name} & \textbf{\#Samples} & \textbf{Min.} & \textbf{Max.} & \textbf{Mean} & \textbf{Min.} & \textbf{Max.} & \textbf{Mean} \\ \hline
Manifold40 & 50 & 6561 & 137055 & 56943.2 & 13134 & 160004 & 101312.9 \\ 
ObjaVerse & 50 & 64933 & 1122245 & 234601.5 & 130088 & 182460 & 159874.9 \\ 
Meshy & 20 & 9058 & 37273 & 20981.5 & 18086 & 74512 & 41924.8 \\ 
\bottomrule
\end{tabular}
\vspace{-6pt}
\caption{Dataset Statistics.}
\label{tab:dataset_stats}
\end{table}



\begin{table*}[htbp]
\centering
\small 
\setlength{\tabcolsep}{4pt} 
\begin{tabular}{l|l|ccc|ccc}
\toprule
\multirow{2}{*}{Dataset} & \multirow{2}{*}{Method} & \multicolumn{3}{c|}{Watermark Quality} & \multicolumn{3}{c}{Asset Utility} \\ 
& & Visibility $\uparrow$ & Placement $\uparrow$ & Readability $\uparrow$ & Geometry $\uparrow$ & Semantics $\uparrow$ & Saliency $\uparrow$ \\ \hline
\multirow{2}{*}{Untextured} & Li et al. & 0.387 & 0.066 & 0.208 & 0.660 & 1.0 & 0.566 \\ 
                           & Ours          & \textbf{0.959} & \textbf{0.793} & \textbf{0.963} & \textbf{1.0} & \textbf{1.0} & \textbf{0.793} \\ \hline
\multirow{2}{*}{Textured}      & Li et al. & 0.462 & 0.076 & 0.295 & 0.591 & 0.984 & 0.515 \\ 
                           & Ours          & \textbf{0.818} & \textbf{0.709} & \textbf{0.822} & \textbf{0.984} & \textbf{1.0} & \textbf{0.814} \\
\bottomrule
\end{tabular}
\vspace{-6pt}
\caption{User study results over GenAI Meshy dataset watermarked with our method and the baseline.}
\label{tab:user_study}
\end{table*}

\subsection{User Study}
 {In order to subjectively analyze the performance of our method compared to the baseline, we conducted a user study involving 10 volunteer participants. Each participant was randomly presented with either a textured or untextured 3D object from the GenAI Meshy dataset, watermarked using our method or the baseline. Participants were then asked to answer the following six "yes/no" questions assessing the watermark quality and utility of the displayed 3D object:}
\begin{itemize}
    \item Are the watermarks \textbf{visible} from different views?
    \item Are the watermarks' \textbf{placement} and orientation good?
    \item Are the watermark texts \textbf{readable}?
    \item Is the asset's \textbf{geometry}/shape preserved? 
    \item Is the asset's \textbf{semantics} preserved?
    \item Are the asset's \textbf{salient} areas protected?
\end{itemize}

In total, 373 data samples were collected, where a value of 1 and 0 were respectively assigned to the "yes" and "no" answers. {The averaged numerical results across all samples are summarized in \cref{tab:user_study}. As shown in the table, the users gave significantly higher scores to our method for both textured and untextured objects in terms of the visibility of the watermarks from multiple views (Visibility), placement and orientation (Placement), and textual readability (Readability) of the watermarks. Specifically, across textured and untextured cases, the baseline scored approximately 46\%, 68\%, and 64\% lower than our method for placement, readability, and visibility, respectively.}

On the other hand, for asset utility, users rated our and the baseline method similarly in terms of preserving the overall semantics and context (Semantics) of the asset after watermarking. However, our method demonstrated superior performance compared to the baseline in preserving the geometry (Geometry) and salient features (Saliency) of the asset, achieving approximately 37\% and 0.26\% higher scores, respectively.

Overall, users generally rated our method slightly higher for watermark quality on untextured objects compared to our method on textured ones. This discrepancy is often due to texture (i.e., color information), which can significantly influence the visibility and readability of watermarks, particularly when the watermark color closely matches the asset's texture. We addressed this issue as a limitation of our method in \cref{sec:limitation}, highlighting its importance for future improvements.


\begin{figure}
  \centering
   \includegraphics[width=1.0\linewidth]{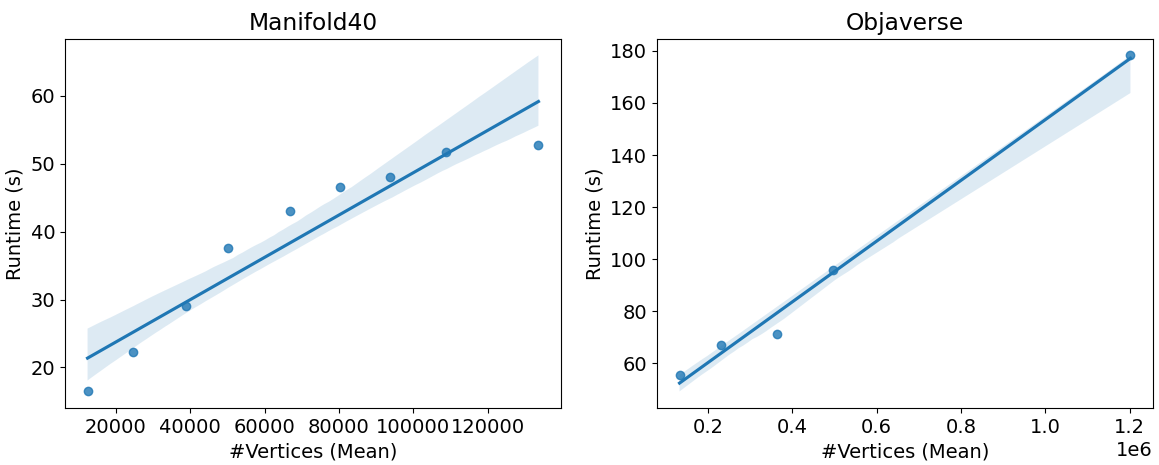}
   \caption{Average runtime (x-axis) required for watermarking models having an average number of vertices (y-axis) for Manifold40 (left) and ObjaVerse (right) datasets,
respectively.}
   \label{fig:run_time_analysis}
\end{figure}

\subsection{Runtime Analysis}
In this section, we provide a runtime analysis of our method. We count all the time required for end-to-end watermarking of an asset, including any preprocessing time, candidate generation, optimization, filtering, and embossing time. The plots for the average runtime (in seconds) corresponding to the average number of vertices are reported in \cref{fig:run_time_analysis} for the Manifold40 and ObjaVerse datasets, respectively. Based on empirical analysis, the overall runtime grows linearly with the number of vertices of the target model. As seen in the plots, a model of 60K vertices requires $\approx$ 30s, and a model of 1.2M vertices requires $\approx$ 180s for watermarking.

\begin{figure}
  \centering
\includegraphics[width=1.0\linewidth]{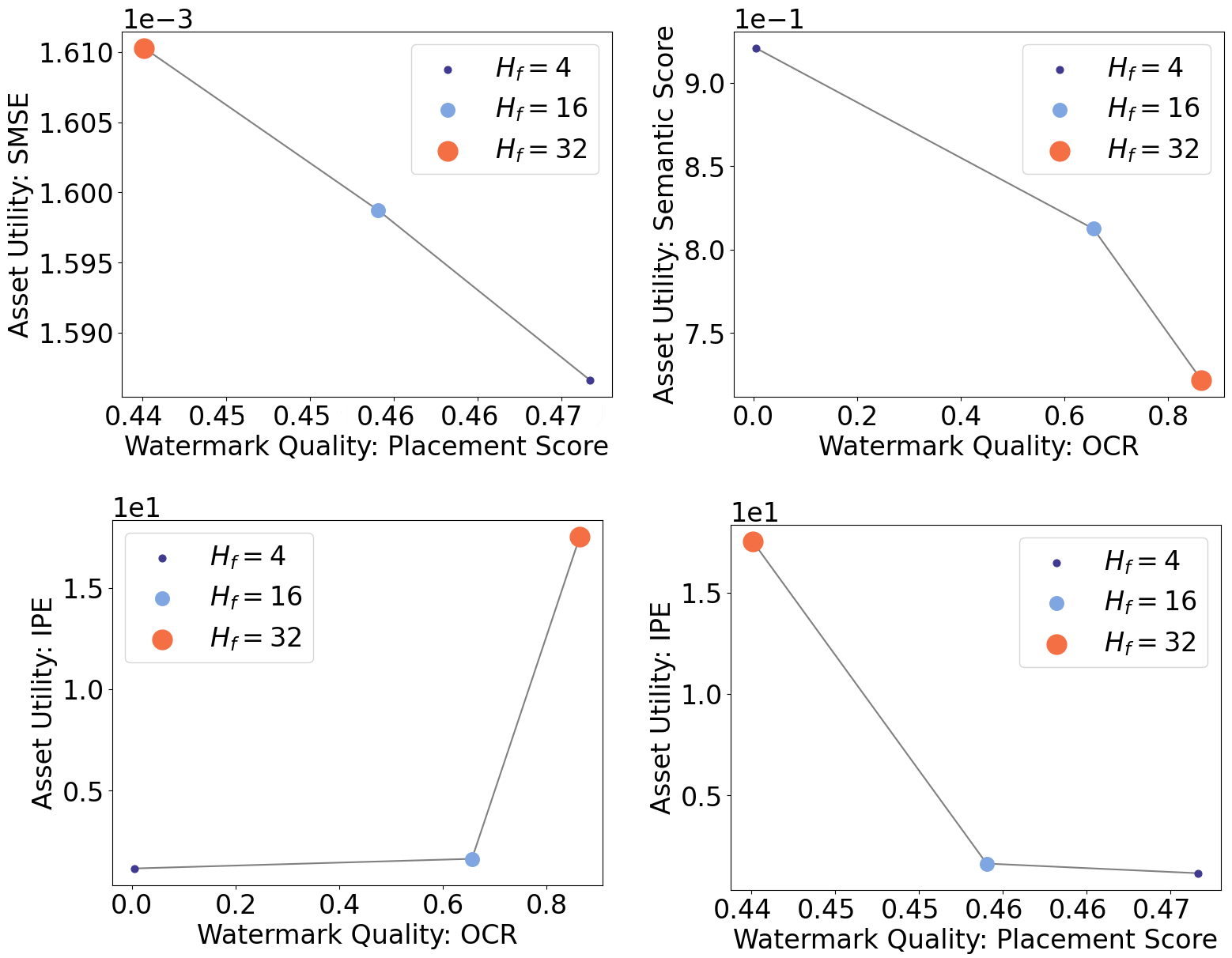}
   \caption{{More trade-off results between the watermark quality and asset quality metrics
on Manifold40. $H_f$ : number of watermarks.}}
   \label{fig:tradeoff_supp}
\end{figure}

\subsection{Watermark Quality vs. Asset Utility Trade-off}
In Sec. 5.1, we studied the trade-off between the watermark quality and asset utility by performing experiments with different numbers of watermarks $H_f = \{4, 16, 32\}$. {Two} trade-off curves including Semantic vs. and Placement, and SMSE vs. OCR were illustrated.

In this section, {four} more trade-off curves in terms of SMSE vs. Placement, Semantic vs. OCR, IPE vs. OCR, and IPE vs. Placement are shown in \cref{fig:tradeoff_supp}. Similar to the trade-off curves provided in the main body of the paper, increasing the number of watermarks results in higher watermark quality, but lower asset utility. However, our method achieves significantly improved trade-off results compared to the baseline ones. For example, our method can achieve an OCR score of $\approx$ 0.85, while providing an IPE error of 18.0. On the other hand, the baseline can achieve a much lower OCR score of 0.30 with a higher IPE error of 20.

Additionally, as shown in \cref{fig:tradeoff_supp}, the effect of the number of watermarks on the placement score vs. the geometry-based SMSE error is very minor. In other words, regardless of the number of watermarks, our method can effectively find the optimal locations and orientations to emboss the watermark without damaging the overall geometry of the original asset.

\begin{algorithm*}
\caption{Candidate Box Generation}
\label{alg:candidate_box_generation}
\begin{algorithmic}[1]
\renewcommand{\algorithmicrequire}{\textbf{Inputs:}}
\REQUIRE Target coordinates \( P_C^i \), target normal \( P_N^i \), watermark string \( Z_m \), watermark size \( Z_s \), watermark font \( Z_f \)
\STATE \( V_i^{ws}, F_i^{ws} \leftarrow \texttt{text\_to\_3d}(Z_m, Z_s, Z_f) \)
\STATE \( V_i^{bs}, F_i^{bs} \leftarrow \texttt{oriented\_bounding\_box}(V_i^{ws}, F_i^{ws}) \)
\STATE \( \alpha^i, \beta^i, \gamma^i \leftarrow \texttt{compute\_angles}([0,0,1], P_i^N) \)
\STATE \( R_i \leftarrow \texttt{generate\_rotation\_matrix}(\alpha_i, \beta_i, \gamma_i) \)
\STATE \( T_i \leftarrow \texttt{generate\_translation\_matrix}(P_i^C) \)
\STATE \( V_i \leftarrow T_i \cdot R_i \cdot V_i^{bs} \)
\STATE \textbf{Output:} $V_i$
\end{algorithmic}
\end{algorithm*}

\subsection{Implementation Details}
In this section, we provide more implementation details of our work. As stated in Sec. 3, the task of 3D visible watermarking has three inputs, namely 1) target model, 2) watermark text, and 3) algorithm parameters such as watermark text font, thickness, and size. We fix the input parameters for all our experiments unless stated otherwise. For input (1), since our method does not depend on texture information, we remove texture information from input models and convert them into standard OBJ file format \cite{wavefront/obj}.

However, our method supports watermarking textured objects, which is done by simply replacing the untextured original model with the textured one during the embossing step. We have provided some qualitative results of watermarking textured models in \cref{fig:meshy_examples}. Other than that, to avoid any variability in metrics computation, we stick to untextured models that are scaled to a fixed size of $30$ and centered at origin $(0, 0, 0)$. Additionally, to preserve computational resources, we decimate the models to keep the number of vertices below $80,000$ for generating watermark boxes by our algorithm. However, during the watermark embossing step, we still use the original undecimated model. Further, we always use ``watermark'' as the watermark text (input 2) for all our experiments and use the default text font as provided by the off-the-shelf library \cite{narayanan_codetigerfont23d_2024} for converting text to 3D mesh. 
We use the thickness (distance between the front and back faces of the watermark) of 0.5 and a fixed watermark size (scale of mesh) of 4 in all our experiments unless stated otherwise. Finally, in the embossing module, we use a fixed value of 0.05 as the extrude strength in \cref{alg:curve_matching_fusion}. 

We sample a fixed number of $H_s=300$ points for generating the initial number of candidate boxes. From this initial set, we obtain the final $H$ by rejecting points that are too close (radius $H_r < 1$). The number of final watermarks after filtering $H_f$ is variable for each model, whose average value is $9.97$ for Manifold40 and $8.82$ for ObjaVerse. The value of $J$, which is the number of sampled points for computing the alignment loss (Eq. 9) is fixed to be $179$, including the midpoints. For optimizing the objective in Eq. 8, we run a fixed number $200$ of gradient descent steps and use stopping criteria of mean loss less than $0.005$. 

We used a 12-CPU-core machine with two NVIDIA GeForce GTX 1080 to run our experiments. All our code was implemented in Python 3.10 and the optimization objective including gradient back-propagation was implemented using Pytorch 3D \cite{ravi_accelerating_2020}. We use off-the-shelf 3D libraries to implement many common mesh operations in this work. Note that, our work can handle all 3D models which can be converted to a mesh and which support 3D Boolean operations. We can simply convert the given format into the mesh model to obtain watermark locations and apply Boolean operations in the end to fuse watermarks into the target format.

\begin{figure*}
  \centering
   \includegraphics[width=0.8\linewidth]{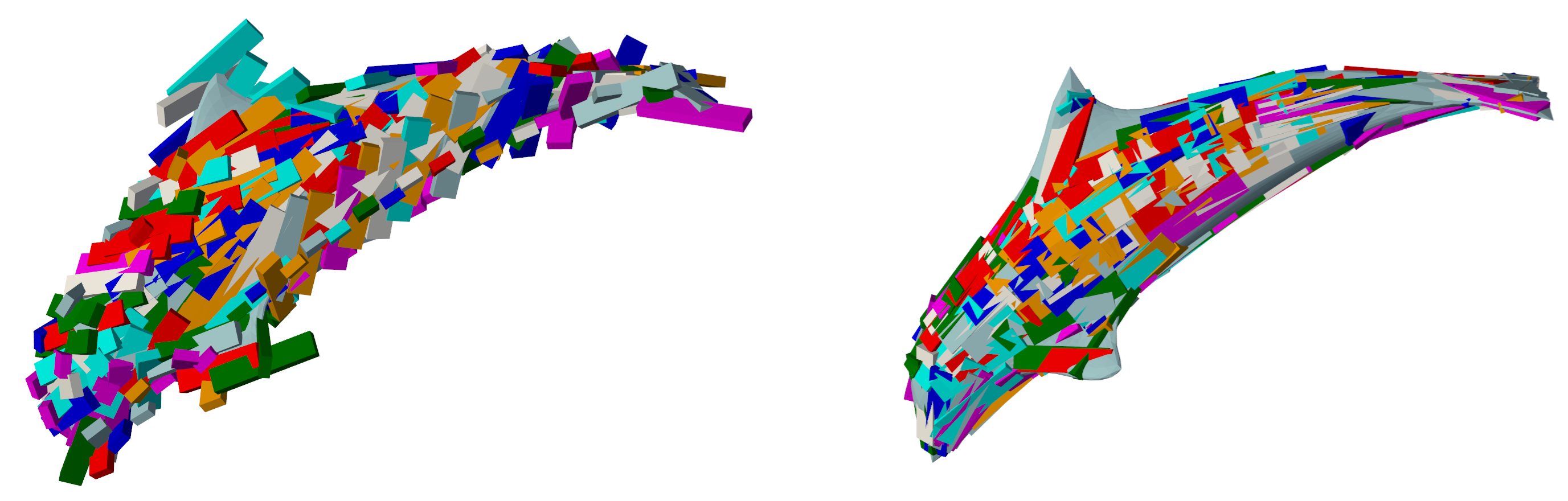}
   \caption{{Effect of optimization on candidate boxes. The left figure shows misaligned candidate boxes placed using \cref{alg:candidate_box_generation} that are fixed (right) by either moving, rotating, or tilting these boxes using the proposed rigid body optimization.}}
   \label{fig:bbox_finetuning_example}
\end{figure*}

\subsection{Initialization (more details)}
In this section, we provide more specific details of the Initialization module (Sec. 4.1). As mentioned earlier, we sample $H$ equidistant points on the surface of the target model. Specifically, we start by randomly sampling $H_s$ points $\{(x, y, z)|x, y, z \in \mathbb{R}\}_{i=1}^{H_s}$ on the surface of the target model. Then, we reject the points that are closer to each other than a radius of $H_r$. After this step, we denote the final set of points, that are approximately equidistant, by $\{{P}_C^i\}_{i=1}^{H}$ and their corresponding surface normal by $\{{P}_N^i\}_{i=1}^{H}$.

Then, for each of these sampled points, we use the procedure in \cref{alg:candidate_box_generation} to generate the candidate boxes. Specifically, we start (Lines 1-2) by generating a watermark mesh $W^s_i (V^{ws}_{i}, F^{ws}_i)$ and its bounding box $B^s_i (V_i^{bs}, F_i^{bs})$ using off-the-shelf algorithms \texttt{text\_to\_3d} and \texttt{oriented\_bounding\_box}. We configure these algorithms to make sure that these meshes are generated at origin $(0, 0, 0)$ and the face of the 3D text faces towards the $+Z$ direction (0, 0, 1), also referred to as front direction. Then, we perform a rigid-body transformation operation (Lines 3-6) to transport the box at the $i^{th}$ sampled location $P_i^C$ and align the box along its normal ${P}^N_i$. Specifically, we use the \texttt{compute\_angles} routine (Line 3) to compute the angles between the front direction of the box $(0, 0, 1)$ and the target direction ${P}^N_i$. We use these angles $\alpha_i, \beta_i, \gamma_i$ to for the rotation $R_i$ (Line 4) and the target location $P_i^C$ to compute the translation matrix $T_i$ (Line 5). Finally, we use these rotation and translation matrices for the final transform (Line 6) to obtain the transformed vertices $V_i$.

\subsection{Finetuning (visualization)}
\label{sec:bounding_filtering}
{
In this section, we present visualizations before and after optimization in the finetuning module (Sec. 4.2). As shown in \cref{fig:bbox_finetuning_example}-left, the boxes placed using the initialization module are misaligned with the surface of the dolphin’s body. After optimization (\cref{fig:bbox_finetuning_example}-right), the boxes' alignment is corrected, and they are positioned accurately to follow the curvature of the dolphin's surface. Specifically, the optimization involves three operations: tilting, rotating, or moving the boxes to achieve proper alignment. For instance, the cyan box situated at the top fin of the dolphin cannot be rotated or tilted and thus needs to be relocated to improve its alignment. Conversely, many boxes on the body can be adjusted by simply rotating or tilting them to correct their alignment.
}

\begin{figure*}
  \centering
   \includegraphics[width=0.8\linewidth]{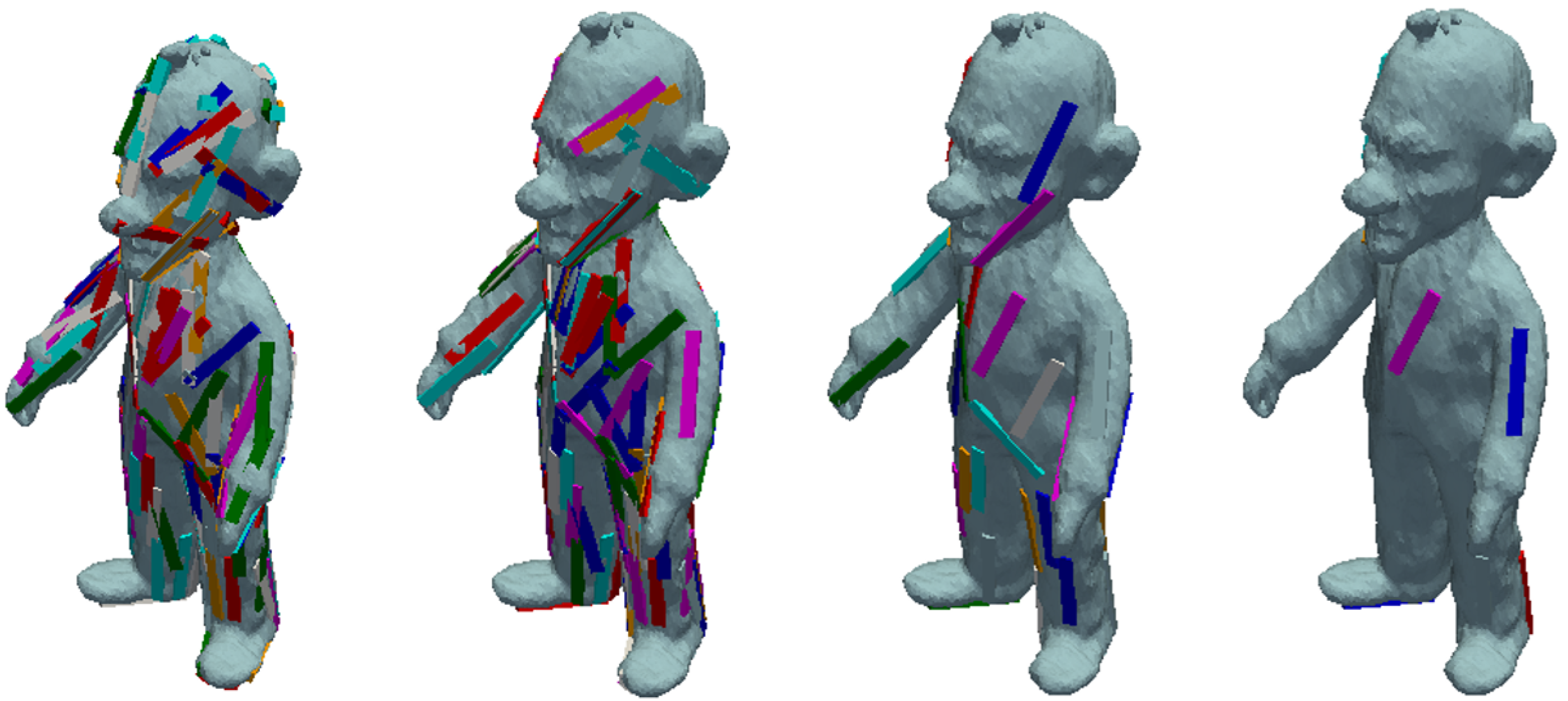}
   \caption{{Step-by-Step Filtering Process: This illustration visualizes the progressive filtering of bounding boxes. From left to right, each image displays the remaining boxes
after applying a specific filter. The first image shows the result after the low roughness score filter. The second image depicts the results after discarding boxes with low loss. The third image presents the outcome of filtering out overlapping and occluded
boxes. Finally, the fourth image displays the final set after applying a multi-octant and multi-angle visibility filter.}}
   \label{fig:filtering_example}
\end{figure*}

\begin{algorithm*}
\caption{Curve Matching Fusion}
\label{alg:curve_matching_fusion}
\begin{algorithmic}[1]
\renewcommand{\algorithmicrequire}{\textbf{Inputs:}}
\REQUIRE original mesh $M$, watermark meshes $\{W_f^i\}_{i=1}^{H_f}$, extrude strength $H_y$
\FOR{$W_i \in \{W_i\}_{i=1}^{H_f}$}
    \STATE $\overline{W}_i \leftarrow \texttt{boolean\_intersection}(M, W_i)$
    \STATE $N^E_i \leftarrow \texttt{closest\_normal}(W_i)$
    \STATE $\tilde{W_i} \leftarrow \texttt{perform\_extrusion}(\overline{W}_i, N^W_i, H_y)$
\ENDFOR
\STATE $M' \leftarrow \texttt{boolean\_union}(M, \{\tilde{W_i}\}_{i=1}^{H_f})$
\STATE \textbf{Output:} $M'$
\end{algorithmic}
\end{algorithm*}

\subsection{Filtering (more details)}
In this section, we provide more details about the individual filtering steps discussed in Sec. 4.3. Going from left to right, \cref{fig:filtering_example} shows the results of various steps of filtering operation. As seen, each filtering step prunes the undesirable boxes and keeps the most important boxes with an aim of high watermark visibility and high asset utility.

\textbf{Well Aligned Boxes with Low Loss:} 
{The loss defined in Eq. 9 quantifies the alignment accuracy of the box with the mesh surface. To reject sub-optimal boxes that are misaligned, we apply straightforward thresholding on the individual box loss. Through observation, we have determined that a loss less than $0.005$ typically indicates well-aligned boxes.}

\textbf{Boxes with Low Local Roughness:} 
{We calculate the local roughness beneath each candidate bounding box and discard boxes exceeding a specific threshold. Here's the detailed procedure. First, we identify all vertices within the $i$-th bounding box \( B_i \). From these vertices, we randomly sample \( H_r \) points and compute the average cross-product of their normals. The local roughness score is defined as the inverse of this average cross-product 
$\mathcal{R}(B_i) = \frac{1}{{H_r}^2} \sum_{j=1}^{H_r} \sum_{k=1}^{H_r} \frac{1}{\cos(N_j^i, N_k^i)}$ where \( (N_j^i, N_k^i) \) are the normals of points inside box \( B_i \). Through analysis, we have determined that a roughness score less than \( 1.25 \) typically indicates boxes located on flatter surfaces.}


\textbf{Non-Overlapping Boxes:} {To handle potential overlaps among candidate boxes, we employ a greedy approach. Initially, we randomly select a box and iteratively discard any box that overlaps with those already chosen. Overlap is determined by checking for intersections among the vertices of the original mesh contained within pairs of bounding boxes.}

\textbf{Non-Occluding Boxes:} {To mitigate potential occlusions of some boxes by parts of the target model, such as under the arms or between the thighs in humanoid models, leading to diminished watermark visibility, we utilize a ray casting method. This approach helps identify and subsequently remove watermarks that are occluded. We sample equidistant points from the front face of each bounding box and cast rays along the normal direction of the watermark. If any of these cast rays intersect with parts of the target model, the watermark is classified as occluded and is subsequently removed.}


\textbf{Multi-Octant Presence:} {To deter model theft, watermarks should be distributed across diverse locations of the model. We achieve this by dividing the model into 8 octants using planes along the \(X\), \(Y\), and \(Z\) axes passing through the model's centroid. Each octant is assigned a watermark. If multiple watermark options exist per octant, we select the one furthest from the watermarks in adjacent octants.}


\textbf{Multi-Angle Visibility:} {In this step, we add extra boxes to ensure the watermark is visible from multiple viewing angles. This prevents attackers from using 2D renders of a 3D object where the watermark might not be visible due to camera angles. Our goal is to position at least one watermark on the visible portion of the model’s surface for each viewing angle. To achieve this, we iterate through fixed angle increments of 30° around the \(X\) and \(Z\) axes and add a watermark if no other existing watermarks are found for that angle.
}


\subsection{Watermark Embossing}
\label{embossing_supplement}
In this section, we provide more details of the novel curve-matching fusion presented in Sec. 4.4. {We start by generating 3D-text watermark meshes \(\{W_i\}_{i=1}^{H_f}\) using a standard text-to-3D algorithm \cite{narayanan_codetigerfont23d_2024}, positioned and oriented according to selected bounding boxes \(\{B_f^i\}_{i=1}^{H_f}\). Then, given the target mesh $M$ and the generated 3D watermarks $\{W_i\}_{i=1}^{H_f}$, we use  \cref{alg:curve_matching_fusion} to obtain the watermarked mesh $M'$. Specifically, first, we apply a boolean intersection operation (Line 2) between the original mesh $M$ and $i$-th watermark mesh $W_i$. Then, we find the extruding normal $N^E_i$ by computing the normal of the closest point on the mesh (Line 3) from the centroid of watermark mesh $W_i$. Next, we perform the extrusion operation (Line 4) of the intersection $\bar{W_i}$ silhouette to give an embossing effect. Finally, we simply take a Boolean union \cite{li_visible_2024} of the extruded watermark meshes $\{\tilde{W_i}\}_{i=1}^{H_f}$ and the original mesh $M$ to obtain the final watermarked result $M'$.

}



\subsection{Evaluation Metrics}
In this section, we provide additional details of the evaluation metrics discussed in Sec. 5.

\subsubsection{Watermark Quality}
\textbf{Watermark Placement Score (WPS):} {WPS measures the alignment between the watermarks and the mesh.} The inputs consist of a mesh $M$ and a bounding box $B_i$. We start by computing vertices of the mesh $M$ that lie inside the bounding box $B_i$ and denote them as $V^{in}$. Then, we find all faces that have at least one vertex in the set $V^{in}$ and denote them as $F^c$. From these faces, we only consider faces that lie completely inside the bounding box (all three vertices in $V^{in}$) and compute their areas. Then, we project these areas in the direction of the front face of the box. Finally, we sum up the areas and divide the sum by the area of the front face of the box to compute the watermark placement score. For multiple bounding boxes, we simply report the mean score {computed across multiple $B_i$s. Note that, we used this approximate procedure to compute area for efficiency purposes as the standard methods do not scale well to a large number of vertices.}

{
\textbf{Ray Casting Visibility (Ray):} {Ray measures the visibility of watermarks from all views of the model.} We begin by generating views of the watermarked mesh \( M' \) by rotating the camera around the \( X \) and \( Z \) axes in 30° increments. For each view \( c_w^t \), we identify candidate watermark meshes oriented within 45 degrees of the camera's direction. Using ray casting, multiple random rays are projected from the top face of each candidate's bounding box towards the camera. A per-watermark score of 1 is assigned if all rays reach the camera without obstruction; otherwise, it is 0. The per-view score is computed by averaging across all per-watermark scores in that view. Finally, the final ray score is obtained by averaging overall per-view scores.
}

\textbf{OCR Visibility (OCR)}: {OCR measures the readability of watermarks from all views of the model.} {We begin by generating renders of the watermarked mesh \( M' \) for each view \( c_w^t \), obtained by rotating the camera around the \( X \) and \( Z \) axes in increments of 30°. Next, we utilize an off-the-shelf OCR detector \cite{keras-ocr_2024} to identify candidate 2D boxes that potentially contain readable text. Subsequently, to account for text orientations that are not left-to-right aligned, we augment the candidate boxes by adding rotations of 90°, 180°, and 270°. Then, we use an off-the-shelf OCR recognizer \cite{keras-ocr_2024} to generate candidate text recognitions. These candidates are then scored using a popular sequence matcher \cite{brancato_could_2020} to quantify their similarity to the ground truth watermark text. Finally, for each view \( c_w^t \), we take the maximum score and average these scores across all views to obtain the final OCR score.}

\subsubsection{Asset Utility}
{
\textbf{Sampled Mean Squared Error (SMSE)}: SMSE aims to compute the Mean Squared Error (MSE) between the watermarked mesh and the original mesh. Since the number of vertices and faces changes after watermarking, it is not possible to compute the MSE directly. Instead, we start by randomly sampling a large number of points on the surface of the watermarked mesh $M'$. Then, we compute the distances of these sampled points from the original mesh M using a standard routine in the Trimesh package \cite{trimesh}. Finally, we report the inverse of the mean distance values as the final SMSE score.

\textbf{Isolated Parts Error (IPE)}: 
{IPE is a measurement of the change in mesh topology before and after watermarking. It is computed as the difference in the total number of isolated parts between the watermarked mesh $M'$ and the original mesh $M$. The motivation is based on the intuition that the number of isolated parts in a model should remain identical after watermarking. An increased IPE captures the cases when a part of the watermark text is disconnected from the model surface. Such isolated parts degrade the asset utility and can be easily removed to damage the watermark message. Lower IPE indicates less change in the mesh topology and therefore better watermark placement.
}

\textbf{Local Curvature Error (LCE)}: {
LCE measures how well the surface curvature of watermarked areas is preserved (as discussed in Sec. 4.4 and \cref{embossing_supplement}). For each vertex on the top face of a watermark, the distance to its nearest neighbor on the original mesh surface $M$ is computed. This process is repeated for every vertex and watermark, and the LCE is calculated as the variance of these distances. A lower LCE indicates that the watermark conforms to the surface curvature, while a higher LCE indicates deviation from the underlying surface curves.}
}

\textbf{Saliency Error (SE)}: {SE is designed to assess if any of the watermark placed covers the salient features of the original mesh.} {It} is computed by first calculating a normalized continuous saliency map of the mesh $M$ using {an off-the-shelf implementation} in \cite{nousias_mesh_2020}. Then, we threshold the saliency map using Otsu’s method \cite{xu_characteristic_2011} to have binary per-vertex salient/non-salient scores. Next, for each watermark bounding box $B_i$, we compute a binary saliency vote for each bounding box indicating if it is placed on a highly salient area. {Specifically, we assign a }value of 1 if the average of thresholded saliency values within the box is greater than 0.5, and a value of 0 otherwise. The average value of saliency votes overall bounding boxes is reported as the saliency score.

\textbf{Semantics Score (SS)}: 
SS is used to measure how well a model's semantics is preserved through measuring the change in visual features after watermarking. To compute it, we start by generating renders of the target \( M \) and watermarked mesh \( M' \) by rotating the camera around the \( X \) and \( Z \) axes in 30° increments. Then, for each pair of corresponding 2D renders, we compute the cosine similarity between their feature vectors extracted using a pretrained ResNet50 \cite{he_deep_2015}. Finally, we average these per-view cosine similarity scores over all views \( \{(c^t_w, c^t_o)\}_{t=1}^T \) (taken at 30° increments around the \( X \) and \( Z \) axes) to obtain the final score.


\begin{figure*}
  \centering
   \includegraphics[width=1.0\linewidth]{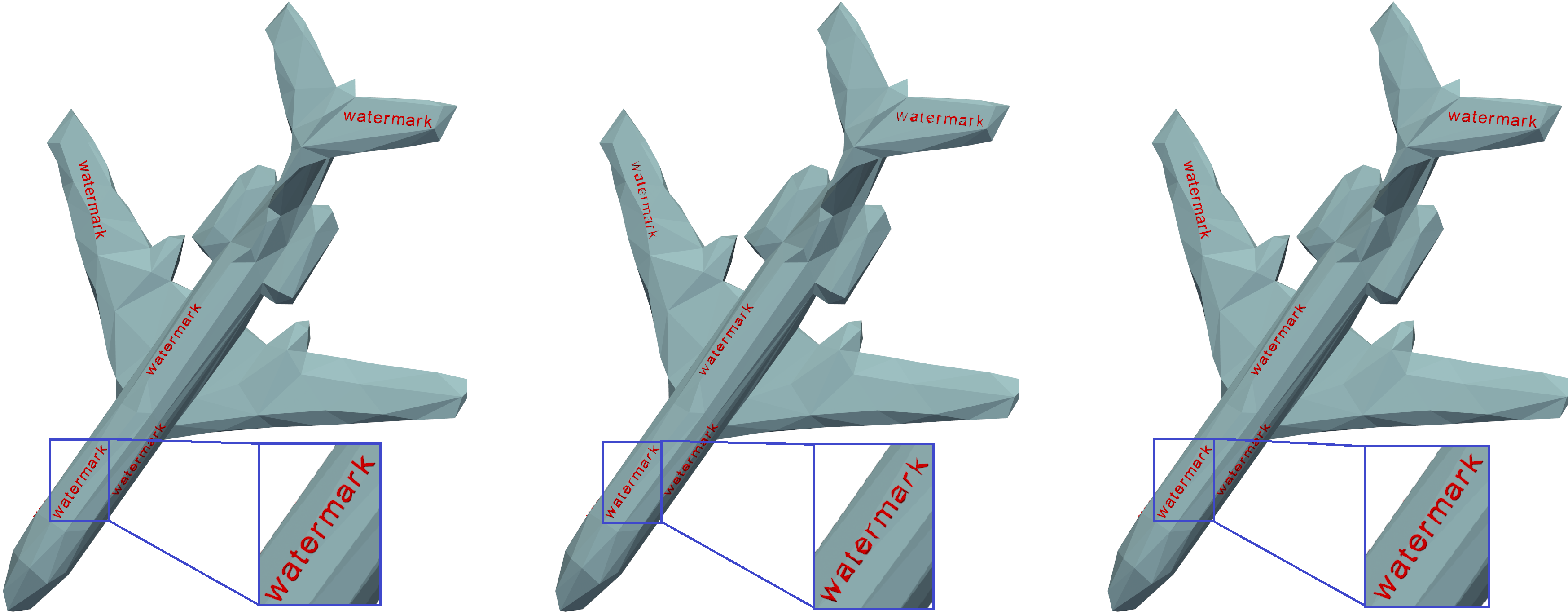}
   \caption{{Impact of mesh editing attacks. From left to right, the first figure shows the original object with no attacks. The second shows the effect of a severe decimation attack (strength = 0.9) and the third shows the effect of a subdivision attack (strength = 2). As seen the watermarks are slightly affected by the decimation attack but they are still visible. On the other hand, the watermarks are unaffected by subdivision attacks.}
   }
   \label{fig:mesh-editing-attacks}
\end{figure*}

\begin{figure*}
  \centering
   \includegraphics[width=1.0\linewidth]{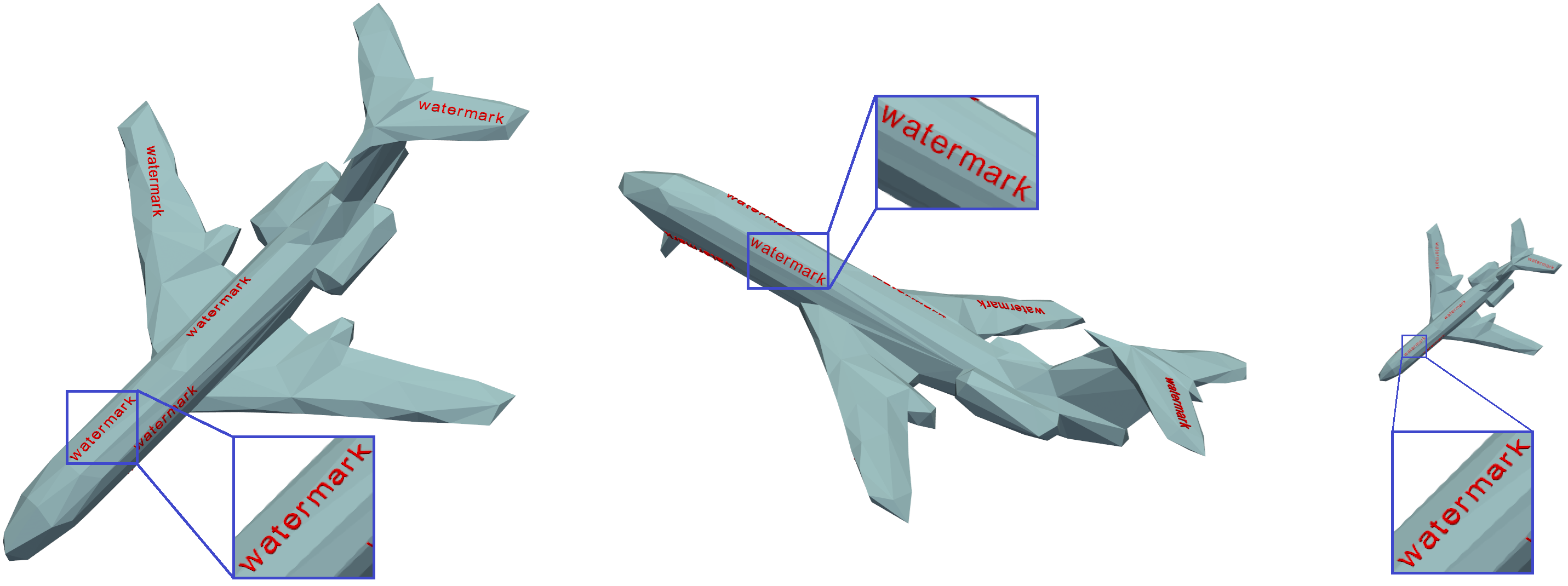}
   \caption{{Impact of geometric attacks. From left to right, the first figure shows the effect of translation to a random position. The second shows the effect of rotation by a random angle and the third shows the effect of scaling the object. As shown, the watermarks move synchronously and are not affected by these inadvertent transformations. }}
   \label{fig:geometric-attacks}
\end{figure*}

\begin{figure*}
  \centering
   \includegraphics[width=0.9\linewidth]{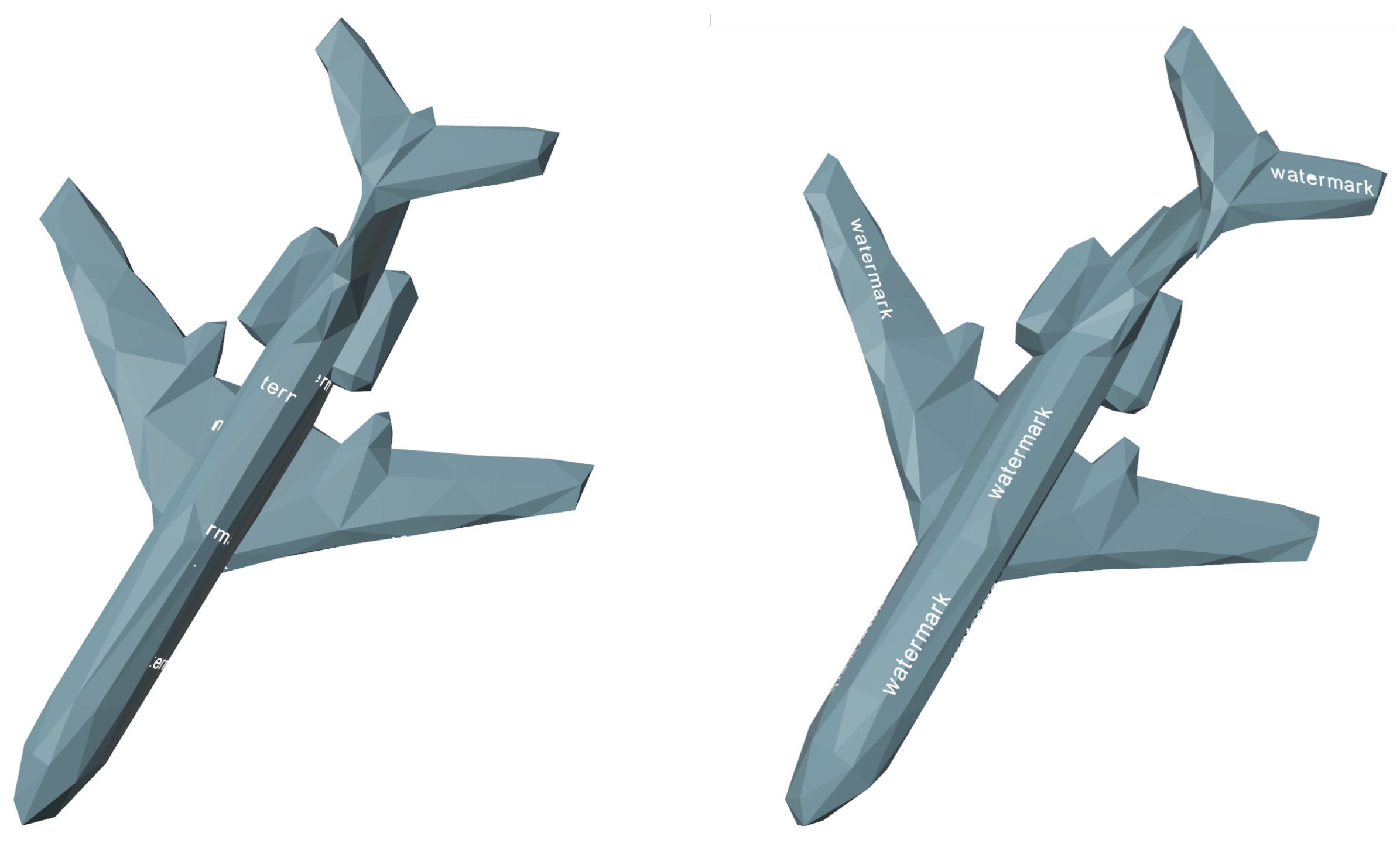}
   \caption{{Impact of Unauthorized Removal Attack: This figure demonstrates the effect of an unauthorized removal attack, where an attacker attempts to eliminate watermarks
by deleting all faces and vertices associated with the watermark mesh. The left side shows the attack applied to Li et al.’s baseline method, where the complete watermark message cannot be read. Conversely, the right side showcases the attack on our proposed method, where we can still easily identify the watermark message, demonstrating its superior resilience compared to the baseline.}}
   \label{fig:removal-attacks}
\end{figure*}

\subsection{More Attacks Analysis}
{
In Sec. 5.4, we presented a preliminary analysis of attacks and robustness. Specifically, we demonstrated the superiority of our method compared to the Li et al. (visible) and Wang et al. (invisible) baselines against three attacks: cropping, unauthorized removal (see Tab. 3 in the main body of the paper), and remeshing attacks (see Fig. 5 in the main body of the paper). In the following section, we extend this analysis and provide additional results.

We begin by analyzing the effects of unauthorized removal attacks on our approach and the Li et al. baseline, as detailed in Sec. 5.4. The qualitative results of this attack are presented in \cref{fig:removal-attacks}. In this attack scenario, we assume a sophisticated adversary who can identify all vertices and faces of the watermarks and remove them using mesh editing software. This task can be quite challenging unless the watermarks are colored with a distinct color. As shown in the figure, even when the attacker knows the vertices and faces, watermarks remain clearly visible in our method due to the silhouette created by the holes. However, for the baseline method, since the watermarks may not fully touch the model surface due to poor orientation, the resulting silhouettes are partial, making the watermarks unreadable.

Next, we conducted tests on typical mesh editing operations such as decimation and simplification using visible watermarks (ours) and invisible watermarks (Wang et al.). We applied a decimation strength of 0.9 and a subdivision strength of 2. These values were chosen to be sufficiently high while ensuring that the visual integrity of the asset remains intact to a normal eye. The results are presented in \cref{fig:mesh-editing-attacks}. In both attacks, the invisible watermark could not be successfully extracted (with less than 50\% bit accuracy), whereas our method preserved the watermark well enough for the message to be clearly readable from multiple angles. Specifically, the clarity of watermarks degraded slightly under the decimation attack but remained completely unaffected by the subdivision attack. We observed that increasing the strength of the decimation attacks could completely erode the watermark, but at that point, the utility of the asset was also significantly degraded.

Lastly, we present qualitative results demonstrating inadvertent geometric operations performed in mesh editing software in \cref{fig:geometric-attacks}. For these operations, such as rotation, scaling, and translation, the watermarks are also transformed synchronously, hence the watermark quality remains unaffected.

}





\subsection{More Qualitative Results}
\label{sec:supp}

\begin{figure*}
  \centering
   \includegraphics[width=1.0\linewidth]{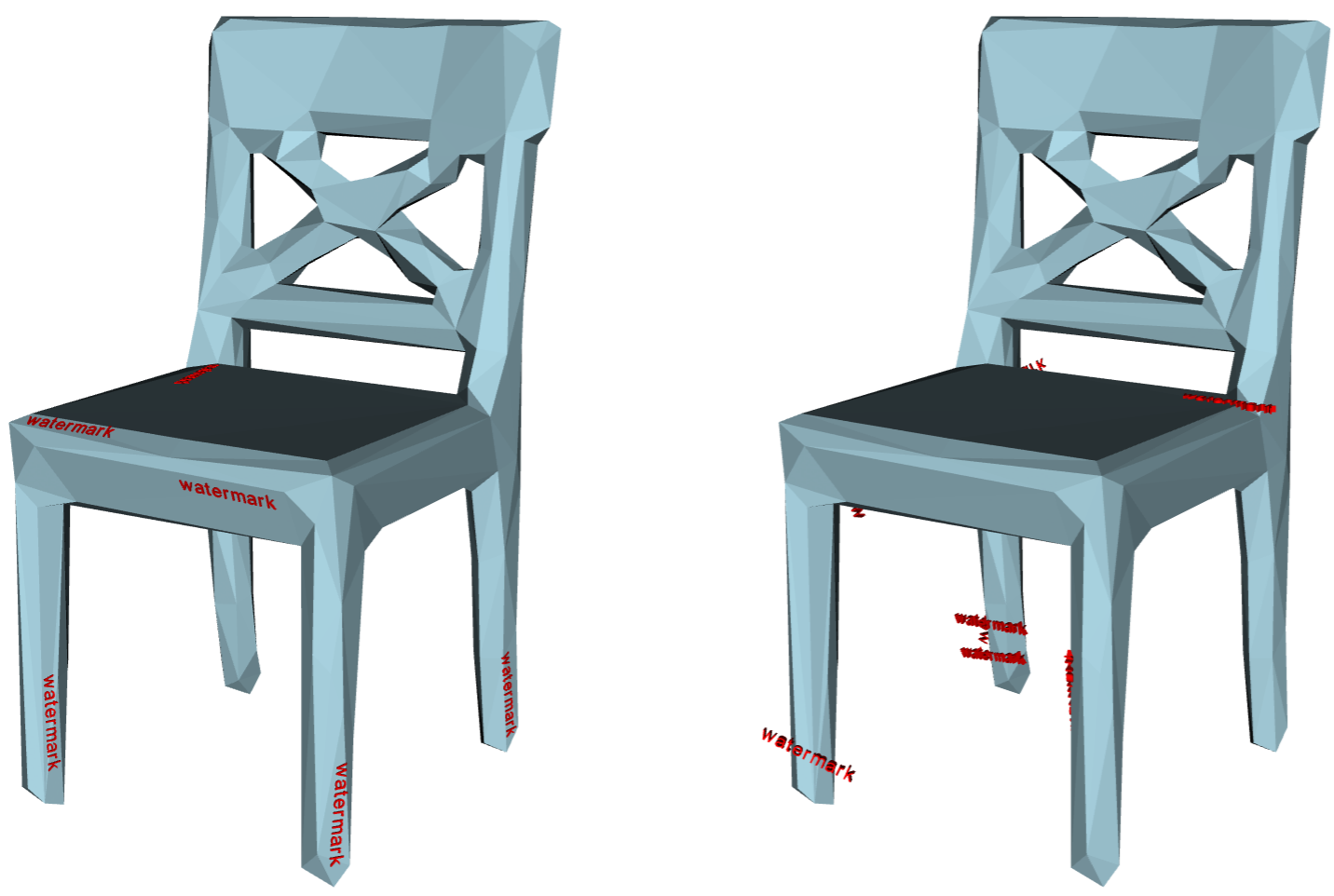}
   \includegraphics[width=1.0\linewidth]{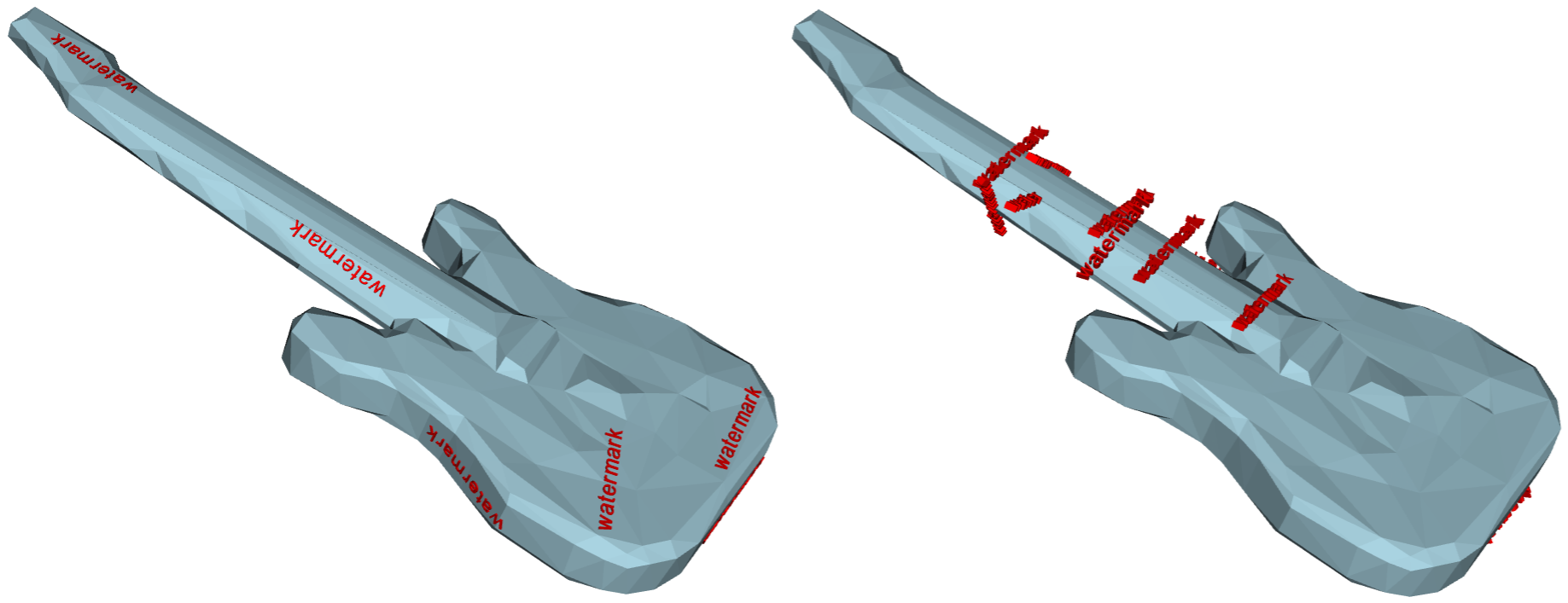}   
   \caption{{Two visual examples from Manifold40 showing 3D models watermarked with our method (left) and Li et al. baseline right). Ours provides better placement quality, readability, and viewability. We colored the watermarks in Red for better observability for the reader.}}
   \label{fig:manifold40_examples}
\end{figure*}

\begin{figure*}
  \centering
   \includegraphics[width=1.0\linewidth]{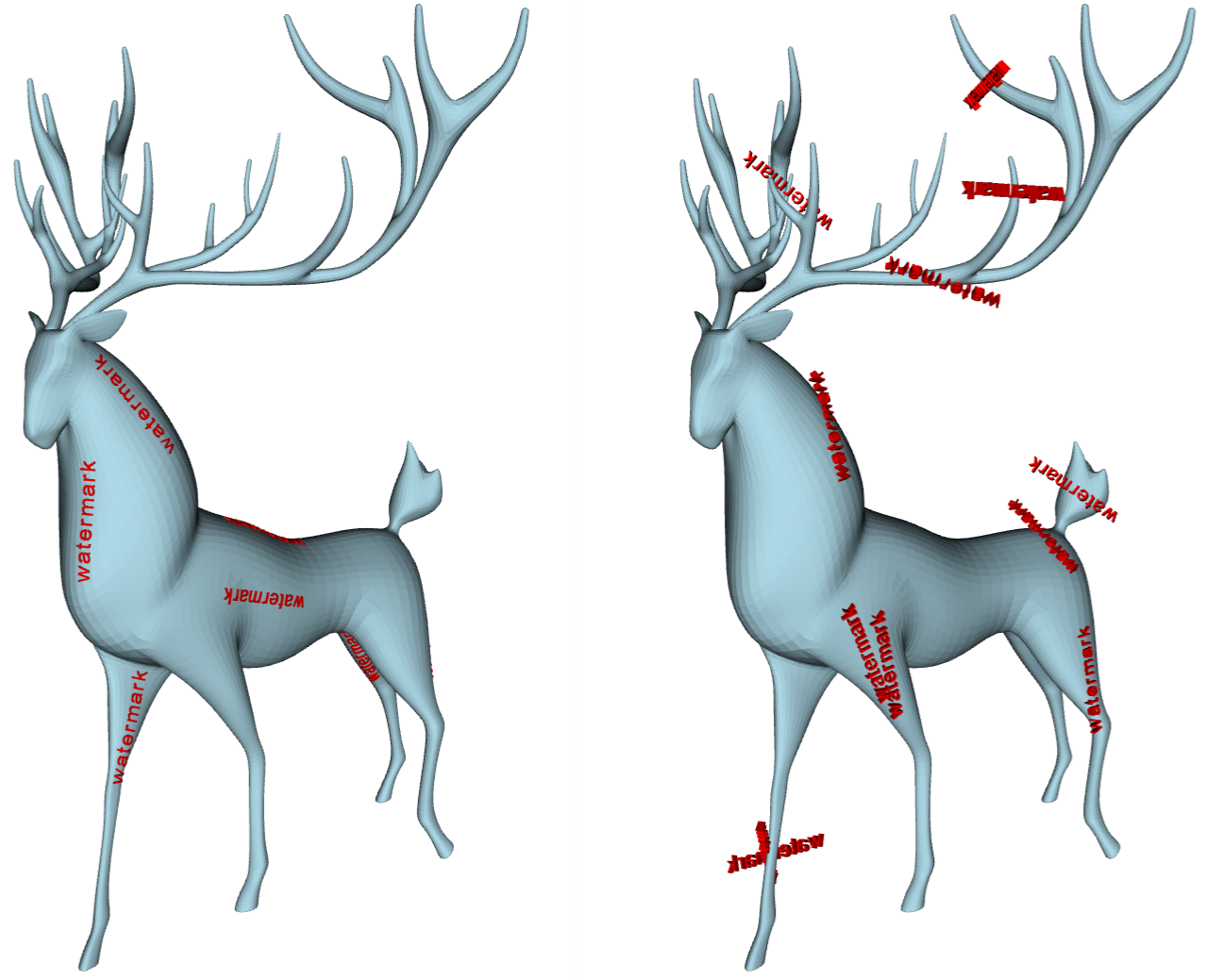}
   \includegraphics[width=1.0\linewidth]{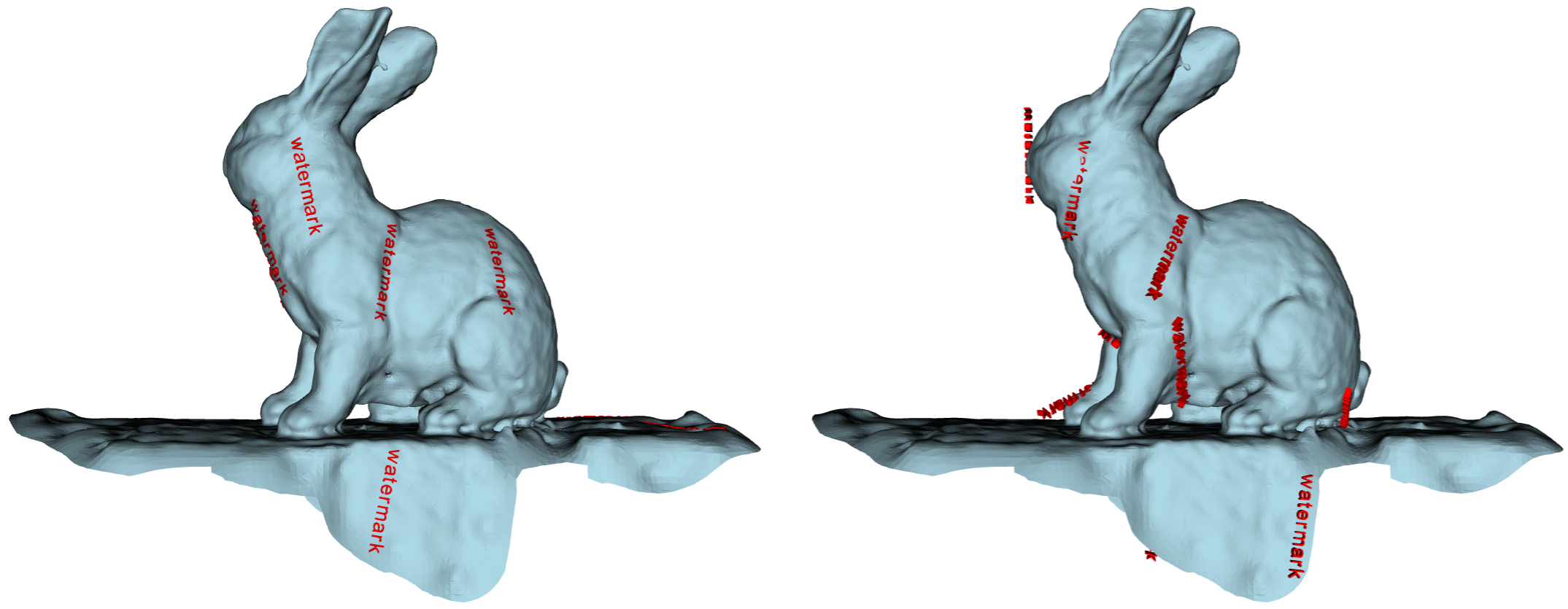}   
   \caption{{Two visual examples from ObjaVerse showing 3D models watermarked with our method (left) and Li et al. baseline (right). Ours provides better placement quality, readability, and viewability. We colored the watermarks in Red for better observability for the reader.}}
   \label{fig:objaverse_examples}
\end{figure*}

\cref{fig:manifold40_examples} and \cref{fig:objaverse_examples}  show some visual examples of the models (from Manifold40 and ObjaVerse) watermarked with our method and the baseline. As shown, compared to the baseline, our method generates watermarks with significantly better placement, orientation, readability, and visibility from multiple views. On the other hand, the baseline produces watermarks that either fly out or are hidden under the surface. 

\begin{figure*}
  \centering
   \includegraphics[width=0.8\linewidth]{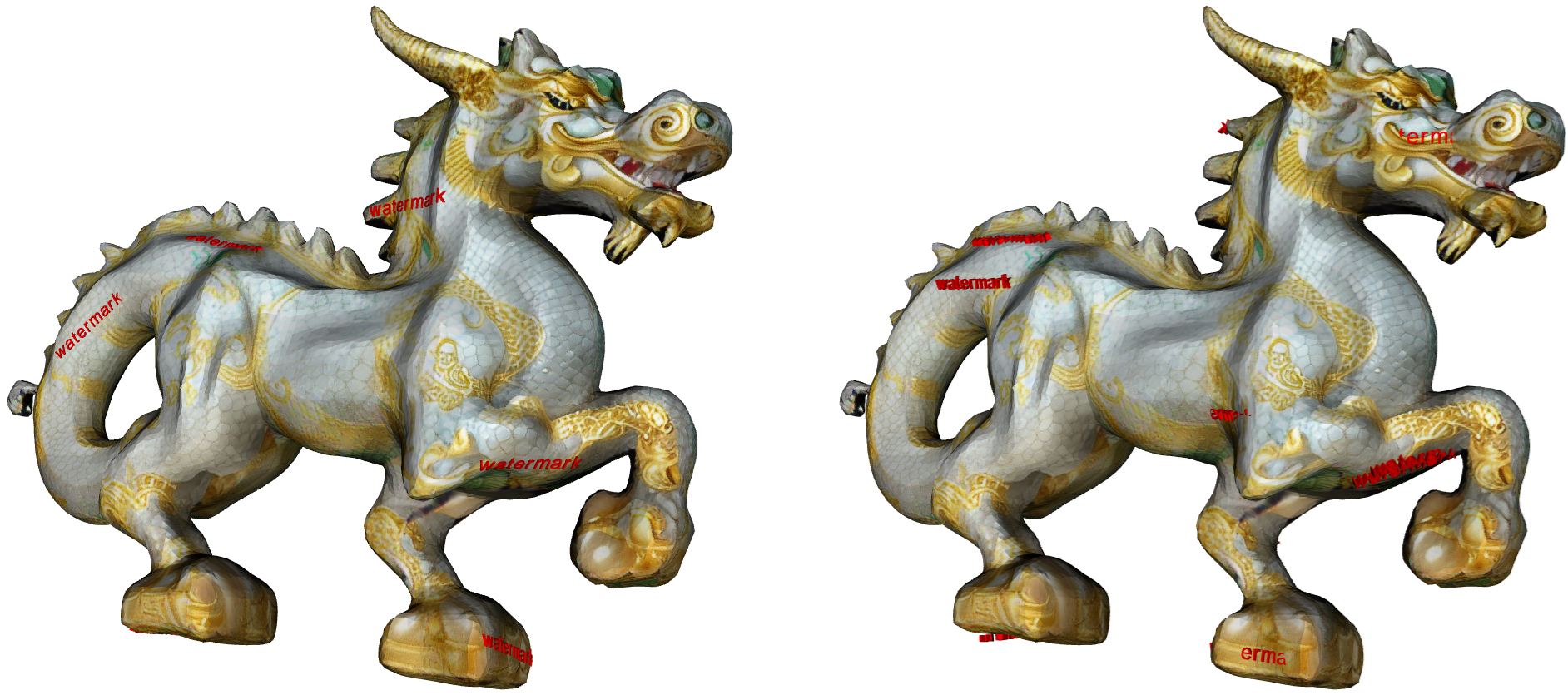}
   \includegraphics[width=0.8\linewidth]{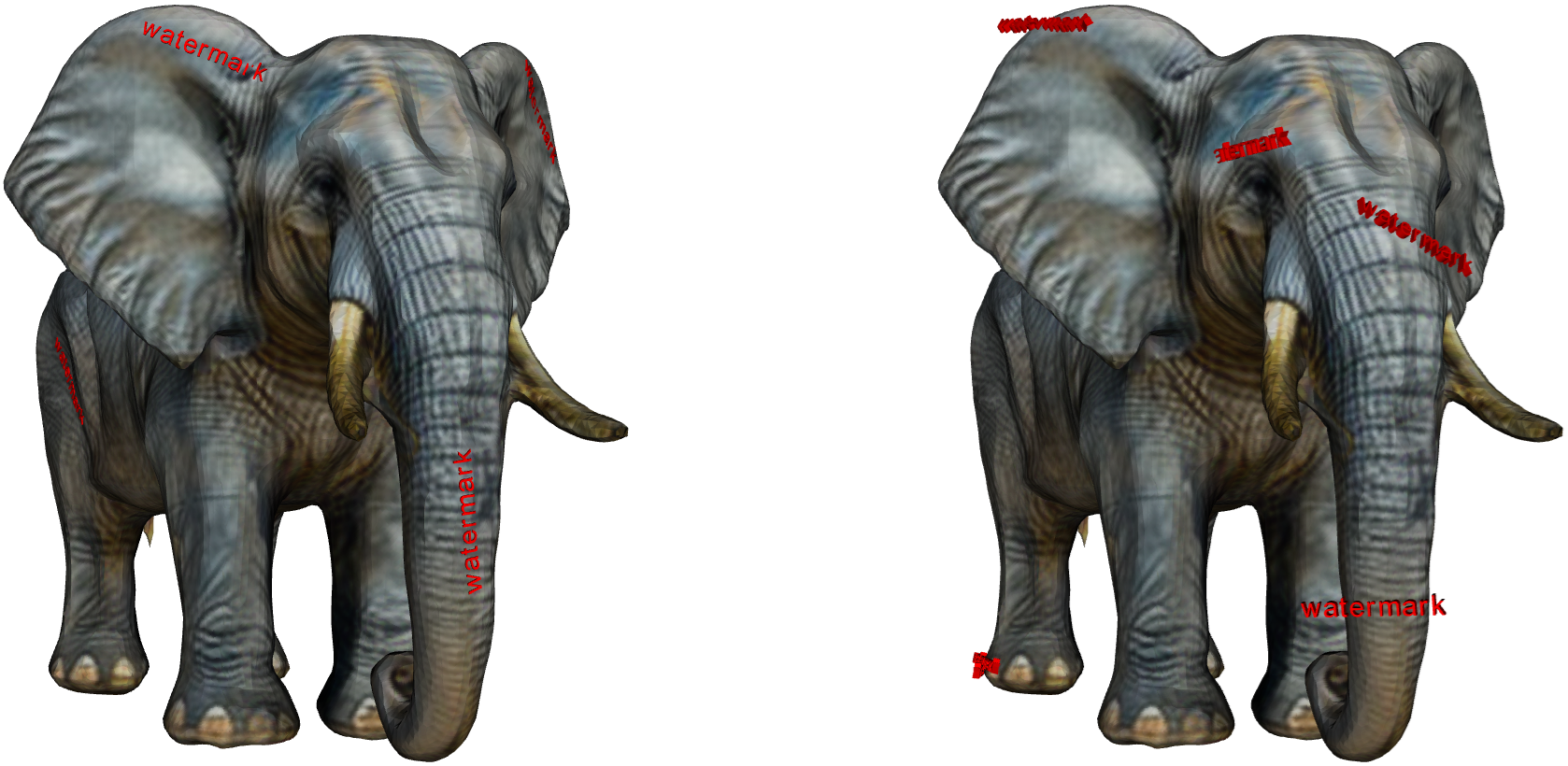}   
   \includegraphics[width=0.8\linewidth]{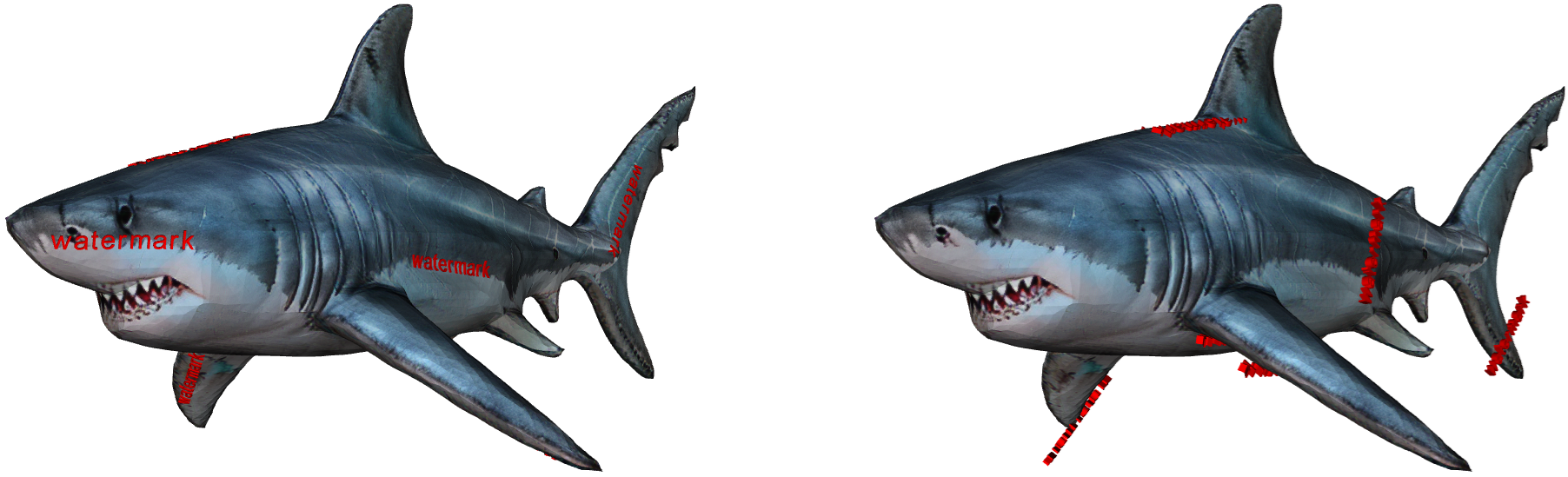}   
   \includegraphics[width=0.8\linewidth]{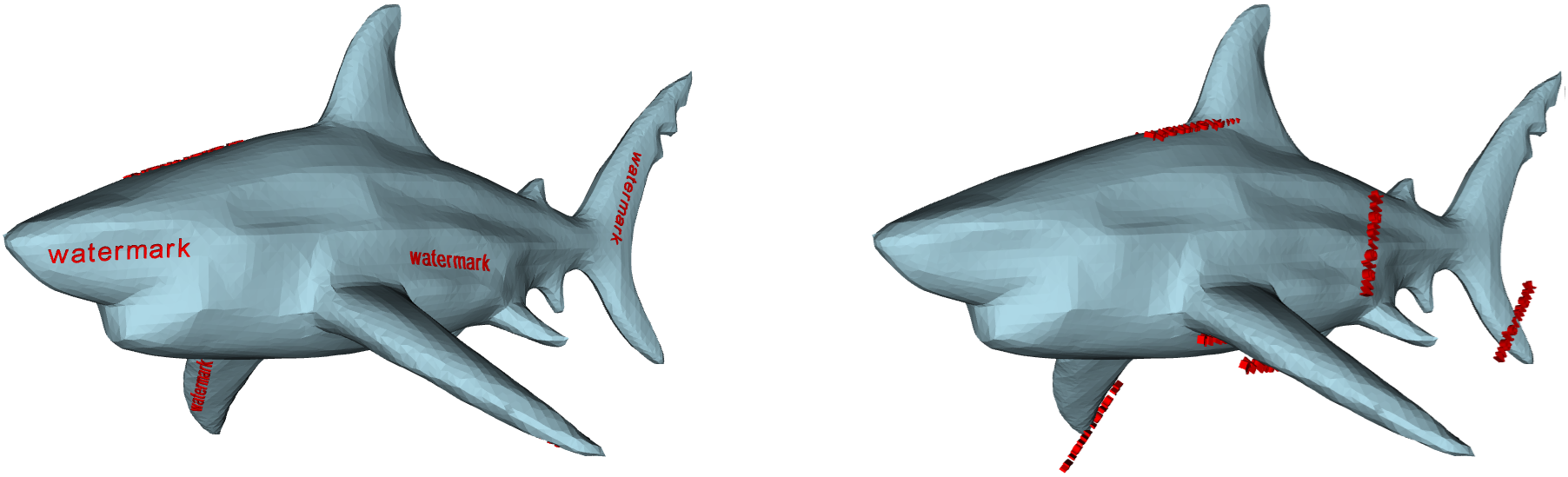} 
   \caption{{Two visual examples from GenAI Meshy showing textured 3D models watermarked
with our method (left) and Li et al. baseline (right). Ours provides better
placement quality, readability, and viewability.}}
   \label{fig:meshy_examples}
\end{figure*}

Please note that we colored (i.e., adding texture) all the watermarks in Red in all the qualitative results for better observability for the reader. However, as also shown in \cref{fig:meshy_examples}, the untextured watermarks still provide high visibility for the IP protection of the objects. 

Moreover, three textured models (from the collected GenAI Meshy dataset) watermarked with our method and the baseline are illustrated in \cref{fig:meshy_examples}. Similar to the visual examples related to human-made datasets in \cref{fig:manifold40_examples} and \cref{fig:objaverse_examples}, our method generates watermarks with significantly better placement, orientation, readability, and visibility from multiple views compared to the baseline. 

It should be noted that our method has been optimized to preserve the most salient features of the mesh without considering the texture information. As a result, for the textured models, it is possible that our method places a watermark on the areas that are visually seen as highly salient due to the presence of the texture (i.e., color information). For example, in the textured shark model in \cref{fig:meshy_examples}, a watermark is placed near the eyes and nose of the shark. However, as also illustrated in the untextured version of the object, such details are not present in the mesh, and the selected area to emboss the watermark is smooth without any salient features.


\subsection{Limitations and Future Work}
\label{sec:limitation}
{
Automated visible watermarking offers a practical framework for several critical scenarios, such as GenAI misuse, merchandise protection, and copyright violation. However, being the first work in this direction, it has several limitations that present significant opportunities for future research.

One significant concern revolves around how well our proposed benchmarks for watermark quality and asset utility align with human perception. The impact of visible watermarks on the perceived asset utility can vary significantly depending on the specific downstream application. Additionally, factors like viewing angle, texture, lighting conditions, and background complexity can influence how watermark quality is perceived in different contexts. This variability and subjectivity complicate the usability and reliability of our proposed metrics across all scenarios universally. Addressing these challenges represents an intriguing direction for future research.
}


Additionally, the robustness of visible watermarks against more intentional attacks poses a significant challenge in certain scenarios. A determined adversary may employ skilled 3D artists to manually remove watermarks and illegitimately sell or use the unwatermarked asset, violating the copyright of the owner. Although such effort will come at a significant cost, due consideration needs to be given to this possibility while employing this technology for practical purposes. Further, it would be an interesting direction for future work to explore and evaluate more sophisticated and automated attacks against our solution and propose a better and a resilient solution. 

Further, our proposed solution works in four independent steps that are not end-to-end optimized for the most efficient performance. Specifically, we believe better performance can be significantly improved by combining the bounding-box fine-tuning and the filtering steps into one single optimization objective. However, this optimization can be challenging due to the various discrete operations in the gradient back-propagation process. We leave it to future work to solve these challenges and propose an improved solution that can further the state-of-the-art in this promising direction. 

Finally, beyond technical considerations, visible 3D watermarking raises practical concerns regarding artistic integrity, where a fine balance needs to be maintained between content protection and user acceptance. Additionally, incorporating 3D visible watermarking in real-time or interactive applications imposes computational overhead that may impact performance and user experience. Addressing these multifaceted and practical challenges is essential for unlocking the full potential of visible 3D watermarking.

\end{document}